\documentclass[twoside]{article}

\usepackage[accepted]{aistats2025}
\usepackage[round]{natbib}

\usepackage[utf8]{inputenc} 
\usepackage[T1]{fontenc}    
\usepackage{hyperref}       
\usepackage{url}            
\usepackage{booktabs}       
\usepackage{amsfonts}       
\usepackage{nicefrac}       
\usepackage{microtype}      
\usepackage{xcolor}         
\usepackage{dsfont}
\usepackage{amsmath}
\usepackage{placeins}
\usepackage{algorithmic}
\usepackage[ruled,vlined,linesnumbered]{algorithm2e}
\usepackage{comment}
\usepackage{enumitem}
\usepackage{float}

\usepackage{subcaption}
\usepackage{amsthm}
\usepackage{amssymb}
\usepackage{mathtools}

\theoremstyle{plain}
\newtheorem{theorem}{Theorem}[section]

\newtheorem{lemma}[theorem]{Lemma}
\newtheorem{corollary}[theorem]{Corollary}
\theoremstyle{definition}
\newtheorem{definition}[theorem]{Definition}
\newtheorem{assumption}[theorem]{Assumption}
\theoremstyle{remark}
\newtheorem{remark}[theorem]{Remark}
\DeclareMathOperator*{\argmin}{argmin}
\DeclareMathOperator*{\argmax}{argmax}

\begin{document}

%

%

\twocolumn[

\aistatstitle{Multi-Player Approaches for Dueling Bandits}

\aistatsauthor{ Or Raveh \And Junya Honda \And  Masashi Sugiyama }

\aistatsaddress{ The University of Tokyo\\ RIKEN AIP \And  Kyoto University\\ RIKEN AIP \And RIKEN AIP\\The University of Tokyo } ]

\begin{abstract}

  Fine-tuning large deep networks with preference-based human feedback has seen promising results. As user numbers grow and tasks shift to complex datasets like images or videos, distributed approaches become essential for efficiently gathering feedback. To address this, we introduce a multiplayer dueling bandit problem, highlighting that exploring non-informative candidate pairs becomes especially challenging in a collaborative environment. We demonstrate that the use of a Follow Your Leader black-box approach matches the asymptotic regret lower bound when utilizing known dueling bandit algorithms as a foundation. Additionally, we propose and analyze a message-passing fully distributed approach with a novel Condorcet-Winner recommendation protocol, resulting in expedited exploration in the nonasymptotic regime which reduces regret. Our experimental comparisons reveal that our multiplayer algorithms surpass single-player benchmark algorithms, underscoring their efficacy in addressing the nuanced challenges of this setting.
\end{abstract}

\section{INTRODUCTION}
\label{sec: Introduction}

Multi-armed bandit (MAB) \citep{auer2002finite} problems, which involve balancing exploration and exploitation to maximize rewards, are key in decision-making under uncertainty. Notable variations of MAB include the dueling-bandit problem and the cooperative multiplayer MAB problem.  In the dueling-bandit scenario \citep{yue2012k}, feedback comes from pairwise comparisons between $K$ arms. This is useful in human-feedback driven tasks such as ranker evaluation \citep{radlinski2008does} and fine-tuning large language models \citep{rafailov2024direct}. These works utilize preference-based optimization, which is based on ideas from dueling bandits or their contextual version \citep{nika2024reward,yan2022human}. Approaches include optimistic algorithms \citep{zoghi2014relative} and algorithms that rely on information theoretic calculations \citep{komiyama2015regret}.
Meanwhile, the cooperative multiplayer MAB focuses on a group of $M$ players collaboratively solving challenges in a distributed environment, enhancing learning through shared information. This is accomplished either by a leader-followers approach \citep{wang2020optimal} or by a fully distributed approach \citep{dubey2020cooperative}. Cooperative multiplayer MAB finds applications in fields like multi-robot systems \citep{lacerda2022decision} and distributed recommender systems \citep{rosaci2009muaddib}.

As training deep networks with human feedback becomes more popular, user numbers grow and tasks shift into complex datasets like images or videos \citep{yang2024using,xu2024imagereward}, which makes distributed approaches important for efficiently gathering feedback. A compelling application arises in fine-tuning large-scale distributed video generation systems \citep{ho2022imagen}, where a network of backend servers directs users to local algorithms that collect preference-based feedback regarding generated video quality. These systems, which must handle high-demand video file communication, optimize performance by relying on low-latency local interactions with nearby servers. Similarly, large-scale distributed ranker evaluation systems \citep{eirinaki2018recommender} also benefit from collaboratively collecting preference-based information. Another example includes a network of self-ordering machines in regional fast-food chains, where each machine collects customer preferences for food products \citep{arsat2023fast}. Local communication is beneficial due to varied preferences across different areas.


To address this, we introduce the $M$-player $K$-arm cooperative dueling bandit problem. This setting, unlike its multiplayer MAB counterpart, demands complex coordination for exploration-exploitation across pairs of arms considering the quadratic number of available comparisons. In multiplayer MAB, communication delays can lead to drawing suboptimal arms, but these still provide useful information regarding the optimal one that reduces future regret. In contrast, in multiplayer dueling bandits, communication delays can lead to pulling pairs of either identical or non-identical suboptimal arms. This not only results in immediate regret but also provides no new information in the former case, and limited information in the latter compared to pulling a pair of optimal and suboptimal arms. Thus, a careful communication strategy that addresses the unique challenges of the multiplayer dueling bandit setting is crucial to achieving low-regret algorithms.

In this study, our focus centers on the widely explored Condorcet Winner (CW) assumption, in which there exists a unique preferable arm \citep{urvoy2013generic}. We introduce a Follow Your Leader (FYL) black-box algorithm, that aligns well with existing dueling-bandit algorithms like Relative Upper Confidence Bound (RUCB) \citep{zoghi2014relative} and Relative Minimal Empirical Divergence (RMED) \citep{komiyama2015regret}, and offers a more natural fit than multiplayer MAB variants \citep{landgren2018social}. Recognizing the limitations of relying on one leader for exploration \citep{yang2023cooperative}, we also propose a decentralized extension to RUCB. Unlike in an MAB counterpart \citep{dubey2020cooperative}, we demonstrate that incorporating an additional CW recommendation protocol between players significantly accelerates the identification of the CW in the nonasymptotic regime. We validate our approach with real-world data experiments. To summarize, our contributions are:
\begin{itemize}
    \item We propose an intuitive black-box algorithm that can be integrated with existing single-player dueling bandit algorithms, conduct comprehensive regret analysis, and show it matches an established asymptotic lower bound when initialized properly.
    \item We conduct an in-depth regret analysis of a fully decentralized multiplayer algorithm based on RUCB, showing asymptotic optimality of the incurred regret up to constants.
    \item We demonstrate that a novel CW recommendation protocol results in a quick identification of the CW in the nonasymptotic regime.  
    \item Through experiments on real data, we show that our algorithms exhibit superior performance compared to a single-player dueling bandit system. 
\end{itemize}

The remainder of this paper is structured as follows: Section~\ref{sec: Problem Formulation} introduces the problem and Section~\ref{sec:Lower Bound} discusses a regret lower bound. In Sections~\ref{sec: Followers Your Leader Black Box Algorithm} and \ref{sec: A Fully Distributed Approach} we propose and analyze our algorithms. Section~\ref{sec: Experiments} showcases experimental results, and Section~\ref{sec: Conclusion and  Future Work} concludes the paper. More related works are discussed in Appendix~\ref{Appendix: Related Work}, and Computational and space complexities as well as communication costs are discussed in Appendix~\ref{Appendix 0A: Computational and Space Complexities}.

\section{PROBLEM FORMULATION}
\label{sec: Problem Formulation}
For some matrix $W$, $w_{ij}$ will denote the $(i,j)$-th element. In the $M$-player $K$-arm dueling bandit problem, each player $m \in [M]=\{1,2,\ldots, M\}$ engages in the environment at each round by drawing an arm pair $\left(i_{m}(t),j_{m}(t)\right)\in [K]^{2}$ and receiving a Bernoulli reward feedback $r_{i_{m}(t),j_{m}(t)}=1-r_{j_{m}(t),i_{m}(t)}$, where $\mathbb{E}r_{i_{m}(t),j_{m}(t)} = q_{i_{m}(t),j_{m}(t)}$ for preference matrix $Q \in \mathbb{R}^{K\times K}$. Here, $\mathbb{E}X$ stands for the expected value of some random variable X. $Q$ is a matrix unknown to players with $q_{ii}=0.5$ and $q_{ij}+q_{ji}=1$ for all $i,j\in [K]$. For brevity, we will use $r_{m}(t)$ when the context is clear. We assume there exists a Condorcet Winner (CW), which is an arm $i$ such that $q_{ij}>0.5$ for all $i\neq j$, and without loss of generality assume that the CW is arm $1$. We also denote by $\mathcal{Q}_{\mathrm{CW}}$ the class of preference matrices with a CW. The assumption regarding the existence of a CW remains widely adopted today. Notably, in Section~\ref{sec: Followers Your Leader Black Box Algorithm} we utilize the RUCB algorithm \citep{zoghi2014relative} for single-player dueling bandits with a CW, which is based on the UCB approach for MAB, and achieves an order optimal $O(K\log T)$ regret with high probability. Additionally, the RMED algorithm \citep{komiyama2015regret} for the same setting has several versions which all employ an empirical divergence minimization approach. In Section~\ref{sec: Followers Your Leader Black Box Algorithm} we use the RMED2FH version which has an $O(K\log T)$ regret and is known to be optimal. 

Players are positioned on a connected undirected communication graph $\mathcal{G}=\left(\mathcal{V},\mathcal{E}\right)$, where $\mathcal{V}=[M]$ denotes the vertices and $\mathcal{E}$ represents the set of edges. We denote $d(m,m')$ as the length of the shortest path between players $m,m'$ on $\mathcal{G}$, and $D=\max_{m,m'}d(m,m')$ as the diameter of the graph. Additionally, the $\gamma$-power of $\mathcal{G}$ for a positive integer $\gamma$, denoted $\mathcal{G}_{\gamma}$, is a graph that shares $\mathcal{G}$'s vertices but includes an edge between any two vertices within a distance of $\gamma$ or less in $\mathcal{G}$. The clique covering number of $\mathcal{G}$ is the minimum number of cliques required to include every vertex. 

Communication operates under a message-passing protocol \citep{dubey2020cooperative,madhushani2021one} with a decay parameter $\gamma$. Each round, player $m$ sends a message $x_{m}^{m'}(t)$ to each player $m'$, which is received after $d(m,m')-1$ rounds if $\gamma > d(m,m')-1$ or is otherwise never obtained. This extends the traditional models of immediate reward sharing where only immediate neighbors communicate ($\gamma=1$) \citep{kolla2018collaborative}, and delayed communication in which all players communicate with each other under delays ($\gamma=D$) \citep{cesa2016delay}. In multiplayer MAB, the two main approaches include distributed decision-making such as message passing UCB \citep{dubey2020cooperative}, and a leader-follower paradigm. Notably, in the latter approach for Bernoulli rewards, \citet{wang2020optimal} introduced a distributed parsimonious exploration strategy with message passing akin to ours called DPE2, in which an elected leader explores using a UCB strategy, and followers exploit a designated arm sent by it. Their approach showcased an optimal regret bound, but required a meticulous derivation that cannot be applied in general.

Let $w_{ij}^{m}(t)=\sum_{\tau=1}^{t}\left(r_{m}(\tau)\overline{\zeta}_{i,j}^{m}(\tau)+(1-r_{m})(\tau)\overline{\zeta}_{j,i}^{m}(\tau)\right)$ denote the number of wins of arm $i$ over arm $j$ experienced by player $m$ until round $t$, where $\overline{\zeta}_{i,j}^{m}(t)=\mathds{1}\{(i_{m}(t),j_{m}(t))=(i,j)\}$ is an indicator for the draw of the ordered pair $(i,j)$. Additionally, $N_{ij}^{m}(t)=\sum_{\tau=1}^{t}(w_{ij}^{m}(\tau)+w_{ji}^{m}(\tau))$ represents the number of visits to $(i,j)$ or $(j,i)$.  
We denote $\Delta_{ij}=q_{ij}-1/2$ as the reward gap, and define the $M$-player cumulative regret as $\mathcal{R}(T) = \sum_{m=1}^{M}\sum_{t=1}^{T}0.5(\Delta_{1i_{m}(t)}+\Delta_{1j_{m}(t)})$. Note that the instantaneous regret for a given player is zero only if the player draws the CW twice in a given round. Additionally, we define $\Delta_{1\max}:=\max\{\Delta_{1i}\}$ and $\Delta_{1\min}:=\min_{i>1}\{\Delta_{1i}\}$.

\section{ASYMPTOTIC LOWER BOUND}
\label{sec:Lower Bound}
We first discuss an asymptotic lower bound for the multiplayer dueling bandit setting in Theorem~\ref{theo:DB Multi-Player Lower Bound}. Let $\mathcal{O}_i=\{j|j \in [K], q_{ij}<1/2\}$ denote the superiors of arm $i$. An algorithm is considered consistent over the class $\mathcal{Q}_{\mathrm{CW}}$ if $\mathbb{E}\mathcal{R}(T) = o(T^{p})$ for all $p>0$ and $Q \in \mathcal{Q}_{\mathrm{CW}}$. Furthermore, we define $N_{ij}(T) = \sum_{m=1}^{M}N_{ij}^{m}(T)$ as the total number of visits to $(i,j)$ or $(j,i)$ in the system, and $\mathrm{KL}(p,q)=p\log\frac{p}{q}+(1-p)\frac{1-p}{1-q}$ as the Kullback-Leibler (KL) divergence between Bernoulli distributions.

\begin{theorem}
\label{theo:DB Multi-Player Lower Bound}
For any consistent algorithm on $\mathcal{Q}_{\mathrm{CW}}$ and $Q \in \mathcal{Q}_{\mathrm{CW}}$, the group regret obeys,
{
\setlength{\abovedisplayskip}{1pt}
\setlength{\belowdisplayskip}{1pt}
\begin{equation*}
    \liminf_{T \rightarrow \infty}\frac{\mathbb{E}\mathcal{R}(T)}{\log T}\geq \sum_{i\in [K]\setminus \{1\}}\min_{j \in \mathcal{O}_{i}}\frac{\Delta_{1i}+\Delta_{1j}}{2\mathrm{KL}(q_{ij},1/2)}. 
    \end{equation*}
}
\end{theorem}
A proof can be found in Appendix~\ref{Appendix A: Lower Bound Proof}. This bound scales as $\Omega(K\log T)$, similar to the single-player case \citep{komiyama2015regret}. The lack of dependence on the number of players, $M$, is typical in multiplayer MAB settings \citep{yang2023cooperative,wang2020optimal} and indicates that, after sufficient time has passed, a multiplayer system can perform as efficiently as a single-player system drawing $M$ pairs per round, regardless of distributed effects. In Sections \ref{sec: Followers Your Leader Black Box Algorithm} and \ref{sec: A Fully Distributed Approach}, we establish matching upper bounds that show this lower bound is tight. However, for a finite horizon $T$, the regret of any algorithm cannot be independent of $M$ \citep{martinez2019decentralized}, so the regret bounds for our algorithms include 
$M$ and $K$ dependent $o(\log T)$ terms. Developing tight lower bounds and corresponding algorithms for these non-leading terms remains an open problem in dueling bandits and multiplayer MAB.

\section{FOLLOW YOUR LEADER BLACK BOX ALGORITHM}
\label{sec: Followers Your Leader Black Box Algorithm}
We now present our first main contribution: a Follow Your Leader multiplayer algorithm capable of employing a single-player base dueling bandit algorithm as a black box. We begin with the following assumption regarding the base algorithm. 

\begin{algorithm}[t]
    \DontPrintSemicolon
    \SetAlgoVlined
    \SetKwInOut{Input}{Input}
    \SetKwInOut{Initialize}{Initialize}
    \caption{Follow Your Leader Black Box (FYLBB)}\label{alg:FYLBlacBox}
    \textbf{Input:} Time horizon $T$, number of arms $K$, number of players $M$, communication graph $\mathcal{G}=(\mathcal{V},\mathcal{E})$, Base algorithm Alg0, Leader Election algorithm LEAlg, Election time $T_{\mathrm{LE}}$.\;
    \textbf{Initialize:} $i(m)=1$ for $m \in \mathcal{V}$.\;
    \For{$t=1,\ldots,T_{\mathrm{LE}}$}{
    \For{$m=1,\ldots,M$}{
        Communicate according to LEAlg.\;
        Draw arms $(c(t),d(t))$ according to Alg0.\;
    }}
    
    \For{$t=T_{\mathrm{LE}}+1,\ldots,T$}{
        \textbf{Leader $m_l$:}\;
        Draw arms $(c(t),d(t))$ according to Alg0.\;
        Get CW candidate $\mathrm{cw}(t)$ from Alg0.\;
        \If{$\mathrm{cw}(t-1) \neq \mathrm{cw}(t)$}{
            Send $\mathrm{cw}(t)$ over the communication graph.\;
        }
        
        \textbf{Followers:}\;
        \For{$m \in [M] \setminus \{m_l\}$}{
            \If{new arm $i$ arrives from leader}{
                $i(m) \leftarrow i$.\;
            }
            Draw pair $(i(m),i(m))$.\;
        }
    }
\end{algorithm}
\begin{assumption}
	\label{assumption: BlackBoxAlg}
     Given a single-player history up to time $t$, the base algorithm outputs:
     \begin{enumerate}[label=(\alph*)]
         \item A pair of arms $(c(t),d(t))$ to draw.
         \item A CW candidate $\mathrm{cw}(t)$, such that $\mathbb{E}\left[\sum_{t=1}^{T}\mathds{1}\{\mathrm{cw}(t)\neq 1\}\right] \leq f(K,T,Q)~.$
     \end{enumerate}
     Furthermore, it has a regret upper bound of $\mathbb{E}\mathcal{R}(T) \leq g(K,T,Q)$, where $f,g$ denote some functions, and $f=o(g)$ with respect to $T$.
 \end{assumption}

Assumption \ref{assumption: BlackBoxAlg} utilizes a structure that several single player dueling bandit algorithms such as RUCB and RMED share. While they do not declare it, these algorithms maintain an internal estimate of the CW, and evaluate the number of rounds in which it is incorrect to be either a constant or $o(\log T)$. They then concentrate on comparing the CW with other arms, thereby reducing the overall regret by a factor of $K$. To maintain generality and avoid algorithm-specific notation, Assumption \ref{assumption: BlackBoxAlg} formalizes this structure using the functions $f,g$ and the estimate $\mathrm{cw}(t)$. For instance, in RUCB $\mathrm{cw}(t)$ can be taken as the hypothesized best arm $\mathcal{B}$ (when exists), and the analysis around the beginning of Theorem 4 in \citet{zoghi2014relative} corresponds to the bound in Assumption \ref{assumption: BlackBoxAlg}(b).

Since we do not assume a leader is given as input, the algorithm starts with a distributed leader-election phase, utilizing some algorithm we denote as LEAlg. This is a well-studied problem with a range of efficient algorithms available \citep{santoro2006design,kutten2015complexity}, typically requiring minimal communication. We concentrate on deterministically identified leader election, where each player is assumed to be initialized with a unique integer identifier and communicates integer values \citep{casteigts2019deterministic}. Utilizing leader-election algorithms without these assumptions is straightforward, but results in a higher communication cost \citep{mitra2022alea}. LEAlg classifies each player as either ``leader'' or ``non-leader'', with only one player holding the former designation, and is completed in $T_{\mathrm{LE}}$ time steps, independent of $T$. An illustrative example of such an algorithm can be found in Appendix~\ref{Appendix B: Follow Your Leader Black Box Algorithm Proofs}, where $T_{\mathrm{LE}}=D+1$. During this initial phase, all players independently draw arms based on the base algorithm. For rounds $t=T_{\mathrm{LE}}+1,\ldots,T$, the elected leader $m_l$ continues to draw arms similarly without receiving feedback from other players, and in rounds where $\mathrm{cw}(t-1) \neq \mathrm{cw}(t)$, initiates a message containing the new $\mathrm{cw}(t)$. This is facilitated through a message-passing protocol with a decay parameter equivalent to the diameter of the communication graph. Simultaneously, each follower $m$ engages in pure exploitation by drawing the arm pair $(i(m), i(m))$, where $i(m)$ is the arm last received from the leader.

While our setup resembles the DPE2 algorithm for multiplayer MAB \citep{wang2020optimal}, Algorithm~\ref{alg:FYLBlacBox} works with any base algorithm, unlike DPE2 which utilizes a specific exploration strategy that requires a meticulous derivation. In addition, unique to the dueling-bandit setting, followers are restricted to pure exploitation of the form $(\mathrm{cw}(t), \mathrm{cw}(t))$, meaning they gain no information about the preference matrix. Thanks to this property, it is relatively easy to adapt single-player dueling bandit algorithms to this framework without the need to carefully adjust the leader's exploration rate, by presenting a candidate arm $\mathrm{cw}(t)$, which is typically a crucial element in the original algorithm. Consequently, our algorithm achieves broader applicability.

\begin{theorem}
	\label{theorem:FYLBlackBox}
     Under Assumption \ref{assumption: BlackBoxAlg}, Algorithm~\ref{alg:FYLBlacBox} satisfies,
     \begin{equation*}
         \begin{aligned}
             \mathbb{E}\mathcal{R}(T)&\leq g(K,T,Q)+M\Delta_{1\max}f(K,T,Q)\\ &+M(T_{\mathrm{LE}}+2D)\Delta_{1\max}
             ~.
         \end{aligned}
     \end{equation*}
 \end{theorem} 
Theorem~\ref{theorem:FYLBlackBox} demonstrates that, in the asymptotic regime, the second and third terms are small compared to the first one, and the regret of Follow Your Leader Black Box (FYLBB) is dominated by the single-player regret bound $g$, thus inheriting the same asymptotic performance. 

\paragraph{Proof Outline} 
The proof involves decomposing the group regret resulting from the leader election phase, the one incurred when followers draw the currently recommended arm, the one when they draw a different arm due to delays, and the leader’s regret itself. We establish that despite the difficulties that occur from the use of a general baseline algorithm and delays, Assumption~\ref{assumption: BlackBoxAlg} and the lack of follower feedback allow us to bound those terms by $f$ and $g$ in a corresponding single-player setting with the base algorithm. This enables us to prove a regret bound despite the generality of this setting. The detailed proof is available in Appendix~\ref{Appendix B: Follow Your Leader Black Box Algorithm Proofs}.

The construction in Theorem~\ref{theorem:FYLBlackBox} enables us to seamlessly adapt single-player dueling bandit results to multiplayer scenarios and even attain asymptotically optimal algorithms.
 \begin{corollary}
	\label{corollary:FYLRMED}
     For any $\epsilon>0$ and $\alpha>1$, the following holds for the regret bound of Algorithm~\ref{alg:FYLBlacBox} with leader election as specified by Algorithm~\ref{alg:Simple Leader Election}.
     \begin{enumerate}[label=(\alph*)]
         \item For RUCB \citep{zoghi2014relative} as the base algorithm,
            {
            \setlength{\abovedisplayskip}{1pt}
            \setlength{\belowdisplayskip}{1pt}
            \begin{equation*}
            \begin{aligned}
                \mathbb{E}\mathcal{R}(T)&=\sum_{i=2}^{K}\frac{4\alpha \log T}{\Delta_{1i}}+O\left(MD+M\frac{K^2 \log K}{\Delta^2_{1\min}}\right)
            \end{aligned}
            \end{equation*}
            }
         \item For RMED2FH \citep{komiyama2015regret} as the base algorithm,
         {
            \setlength{\abovedisplayskip}{1pt}
            \setlength{\belowdisplayskip}{1pt}
         \begin{equation*}
                \begin{aligned}
                    \mathbb{E}\mathcal{R}(T) &=\sum_{i=2}^{K}\min_{j \in \mathcal{O}_{i}}\frac{\Delta_{1i}+\Delta_{1j}}{2\mathrm{KL}(q_{ij},1/2)}\log T\\
                    &+O\left(M(D+K^{2+\epsilon}\log\log T)+\frac{K\log T}{\log\log T}\right)
                \end{aligned}
            \end{equation*}
        }
     \end{enumerate}
 \end{corollary}
By employing RMED2FH as the base algorithm, our approach achieves performance that matches the lower bound from Theorem~\ref{theo:DB Multi-Player Lower Bound} in the asymptotic regime. Utilizing RUCB yields an $O(K\log T)$ algorithm that remains independent of $M$ under these conditions. Notably, for both algorithms, the candidate $\mathrm{cw}(t)$ corresponds to an arm already retained by the original algorithms. Lemma~\ref{lemma: RMED RUCB CW mistake bound} is used to prove Corollary~\ref{corollary:FYLRMED} by providing the bounds $f,g$ for RUCB and RMED2FH. 

While Algorithm~\ref{alg:FYLBlacBox} can leverage any base algorithm satisfying Assumption \ref{assumption: BlackBoxAlg} with comparable asymptotic guarantees, its empirical performance is suboptimal in certain scenarios, as demonstrated in Section~\ref{sec: Experiments}. This arises from the lack of cooperation for exploration, as indicated by the nonasymptotic term $O\left(Mf(K, T, Q)\right)$, representing the regret incurred by the leader in identifying the CW. Moreover, an FYLBB approach is less suitable for systems where players enter and exit dynamically as it complicates leader election. These concerns motivate us to develop an expedited exploration approach in the next section.

\section{A FULLY DISTRIBUTED APPROACH}
\label{sec: A Fully Distributed Approach}
While Algorithm~\ref{alg:FYLBlacBox} offers an intuitive black-box approach, exploration without cooperation can lead to significantly large finite-time regret, motivating us to introduce a distributed strategy. We demonstrate that a novel message-passing extension of RUCB in Algorithm~\ref{alg:MPRUCB} where players share rewards and update the CW candidate based on recommendations from other players, is an effective distributed approach for this setting. Unlike a naive message-passing extension \citep{madhushani2021one} which would modify the RUCB algorithm
\begin{algorithm}[t]
\DontPrintSemicolon
\SetAlgoVlined
\SetKwInOut{Input}{Input}
\SetKwInOut{Initialize}{Initialize}
\caption{Message-Passing RUCB (MP-RUCB)}
\label{alg:MPRUCB}
\textbf{Input:} Time horizon $T$, number of arms $K$, number of players $M$, $\alpha>1.2$, communication graph $\mathcal{G}=(\mathcal{V},\mathcal{E})$, decay parameter $\gamma$.\;
\textbf{Initialize:} $\tilde{W}^{m}:=(\tilde{w}_{ij}^{m}),\tilde{N}^{m}:=(\tilde{N}_{ij}^{m}) \leftarrow 0_{K\times K},\mathcal{B}^{m},\mathcal{B}^{m}_{\mathrm{rec}},x_{m}(t)\leftarrow \emptyset,~\forall~ t\leq 1,~m \in \mathcal{V}$.\;
\For{$t=1,2,\ldots,T$}{
    \For{$m=1,\ldots,M$}{
        $U^{m} := (u_{ij}^{m}) \leftarrow \frac{\tilde{W}^{m}}{\tilde{N}^{m}} + \sqrt{\frac{\alpha \log t}{\tilde{N}^{m}}}$.\; 
        $u_{ii}^{m} \leftarrow 0.5$ \textbf{for} $i \in [K]$.\;
        $\mathcal{C}^{m} \leftarrow \{i \mid u_{ij}^{m} \geq 0.5 \ \forall j \in [K]\}$.\;
        $\mathcal{B}^{m} \leftarrow \mathcal{B}^{m} \cap \mathcal{C}^{m}$.\;
        \textbf{If} $|\mathcal{C}^{m}|=1$ \textbf{then} $\mathcal{B}^{m}\leftarrow \mathcal{C}^{m}$.\;
        \textbf{If} $|\mathcal{B}^{m}_{\mathrm{rec}}\cap\mathcal{C}^{m}|\geq1$ \textbf{then} set $\mathcal{B}^{m}$ to a random element from $\mathcal{B}^{m}_{\mathrm{rec}}\cap\mathcal{C}^{m}$.\;
        \uIf{$\mathcal{C}^{m} = \emptyset$}{
            pick $i_m(t)$ randomly from $[K]$.\;
        }
        \Else{
            \uIf{$\mathcal{B}^{m} \cap \mathcal{C}^{m} \neq \emptyset$}{
                pick $i_m(t)$ from $\mathcal{B}^{m}$.\;
            }
            \Else{
                    pick $i_m(t)$ randomly from $\mathcal{C}^{m}$.\;
            }
        }
        $j_m(t) \leftarrow \argmax_{j} u_{ji}^{m}$.\;
        Draw $(i_m(t), j_m(t))$ and obtain $r_{m}(t)$.\;
        $x_{m}(t) \leftarrow \langle m, t, i_m(t), j_m(t), r_{m}(t) \rangle$.\;
        Send $x_m(t)$ in message-passing.\;
    }
    \For{$m=1,\ldots,M$}{
        $x_{m}^\text{batch}(t) \leftarrow \{x_{m'}(t') \mid d(m,m') = t-t' \leq \gamma - 1\}$.\;
        Update $\tilde{W}^{m}$ with samples in $x_{m}^\text{batch}(t)$.\;
        $\tilde{N}^{m}\leftarrow \tilde{W}^{m}+(\tilde{W}^{m})^{T}$.\;
        $\mathcal{B}^m_{\mathrm{rec}} \leftarrow \{i \mid (i,i) \in x_{m'}(t') \land d(m,m') = t-t' \leq \gamma - 1\}$.\;
    }
}
\end{algorithm}
such that data sent by other players is only utilized to construct improved confidence intervals, in Algorithm~\ref{alg:MPRUCB} players also share CW candidates. This fosters a more effective exploration that results in expedited identification of the CW by players. As a result, compared to a naive message-passing application, the incurred from drawing suboptimal arms is reduced. Detailed proofs for the claims in this section are found in Appendix~\ref{sec:Appendix C: Message Passing RUCB Proofs}.

In lines 5--21 of Algorithm~\ref{alg:MPRUCB}, each player follows a local version of RUCB. Computing the UCB terms $u_{ij}^{m}$ for each arm pair $(i,j)$ is done elementwise in line 5, where we define $x/0=1$ for any number $x$. To perform this calculation, player $m$ maintains the matrix $\tilde{W}^{m}(t):=(\tilde{w}^{m}_{ij}(t))$. Here, $\tilde{w}^{m}_{ij}(t)$ represents information about the number of wins of $i$ against $j$ up to round $t$, known to the player from their own and other players' experiences up to that round. Similarly, $\tilde{N}^{m}(t):=(\tilde{N}^{m}_{ij}(t))$ represents the known number of visits up to round $t$. This is distinct from the local counters $w^{m}_{ij}(t)$ and $N_{ij}^{m}(t)$, which results in each player utilizing information gathered by other players in their decision-making. Additionally, each player maintains its own champion set $\mathcal{C}^{m}$ with arms likely to win against all other arms, a CW candidate set $\mathcal{B}^{m}$ that may either contain one candidate arm or be empty, and a set of recommended arms $\mathcal{B}^{m}_{\mathrm{rec}}$. The CW candidate set $\mathcal{B}^m$ is updated with a random element from $\mathcal{C}^{m}\cap \mathcal{B}^{m}_{\mathrm{rec}}$ whenever this set is non-empty, and is otherwise updated based on $\mathcal{C}^{m}$. In line 18, if there is a tie we draw $j_m(t) \neq i_m(t)$. Players communicate through a message-passing protocol with a decay parameter $\gamma$, as described in Section~\ref{sec: Problem Formulation}. For simplicity, we denote the message sent by player $m$ at round $t$ as $x_{m}(t)$ in line 20, which contains the drawn arms and the sampled reward. In lines 23--26, players receive messages and update $\tilde{W}^{m}$ and $\tilde{N}^{m}$ based on all available arms and reward samples. Moreover, whenever a new message includes an exploitation draw of the same arm twice, that arm is added to the recommended set $\mathcal{B}^{m}_{\mathrm{rec}}$ for use in the next round. 
 
We begin with a lemma that shows with high probability that for rounds larger than
\begin{equation*}
    C(\delta):=\left(4MK^2\left(3+2\log \left(d_{\max}^{\gamma}+1\right)\right)/\delta\right)^{\frac{1}{1.7\alpha -1.4}}~,
 \end{equation*}
visitation numbers to arm pairs other than $(1,1)$ cannot be too large across the system. 

\begin{lemma}
	\label{lemma:MP visitation bound}
	For any $\alpha  > 1.2$, $\delta>0$ and $T_0 \geq C(\delta)$, define $N_{ij}^{m,T_0}(t)$ as the number of visitations of player $m$ to $(i,j)$ between rounds $T_0$ and $t$, and $\overline{\Delta}_{ij}:=\min\{\Delta_{1i},\Delta_{1j}\}$ for $i, j\neq 1$, $\overline{\Delta}_{1i}:=\Delta_{1i}$.
 Additionally, for some $\gamma'\leq \gamma$ let $\mathcal{C}(\mathcal{G}_{\gamma'})$ be the set of all cliques in $\mathcal{G}_{\gamma'}$. Then, with probability $1-\delta$, $\sum_{m \in C}N_{ij}^{m,T_0}(t)\leq \left[4\alpha\log t/\overline{\Delta}_{ij}^2+(\gamma' +2)|C|\right]\mathds{1}\{i\neq j\}$ holds for all rounds $t>T_0$, parameters $\gamma' \leq \gamma$, cliques $C \in \mathcal{C}(\mathcal{G}_{\gamma'})$ and pairs $(i,j)\neq(1,1)~$.
 \end{lemma}
 
The establishment of this bound is pivotal, as it allows us to both bound the leading term of the group regret with a multiplicative factor smaller than $M$, and to bound the non-leading term in $T$ corresponding to the regret incurred before identifying the CW. A naive analysis akin to the single-player case would lead to only bounding the local counters $N_{ij}^{m,T_0}(t)$, resulting in a regret that scales like $O(MK\log T)$. To avoid that, we capitalize on the shared observations among players to limit the total number of visits within any clique, and rely on the fact that the first arm is drawn from the player-specific champion set regardless of recommendations. Unlike existing multiplayer MAB methods \citep{madhushani2021one}, we demonstrate the visitation counter bound holds with high probability even for rounds in which the pair $(i,j)$ is not drawn by any player, and for any clique in the graph to facilitate faster CW identification.

Before presenting the group regret bound for Algorithm~\ref{alg:MPRUCB}, we introduce the following term.
 \begin{equation*}
         \tilde{C} = \frac{1.7\alpha -2.4}{1.7\alpha -1.4}\left(4MK^2\left(3+2\log \left(d_{\max}^{\gamma}+1\right)\right)\right)^{\frac{1}{1.7\alpha-1.4}}.
     \end{equation*}

 \begin{theorem}
	\label{theorem: MP}
     For any $\delta>0$ and $\gamma'\leq \gamma$, define $\mathcal{D}:=\sum_{i<j}4\alpha/\overline{\Delta}_{ij}^2$, $\Gamma(m,\gamma')$ as the size of the largest clique player $m$ belongs to in $\mathcal{G}_{\gamma'}$ and
     \begin{equation*}
     \begin{aligned}
         \hat{C}: &= \Delta_{1\max}\sum_{m=1}^{M}\min_{\gamma'\leq \gamma}\left(K^2(\gamma'+2)+\frac{2\mathcal{D}}{\Gamma(m,\gamma')}\log 2\mathcal{D}\right)\\
         &+(2\tilde{C}+K(3\gamma+2))M\Delta_{1\max}~.
     \end{aligned}
     \end{equation*}
     Let $\chi(\mathcal{G}_{\gamma})$ denote the clique covering number of $\mathcal{G}_{\gamma}$. Then for any $\alpha > 1.4$ and all times $T$,
        \begin{equation*}
        \begin{aligned}
            \mathbb{E}\mathcal{R}(T) &\leq \sum_{j>1}\frac{2\alpha \chi(\mathcal{G}_{\gamma})}{\Delta_{1j}}\log T+\hat{C}~.
        \end{aligned}
        \end{equation*}
 \end{theorem}
The regret bound presented in Theorem~\ref{theorem: MP} demonstrates that a message-passing protocol is an effective distributed approach for the dueling bandit problem. It emphasizes the significance of leveraging candidate recommendations in reducing the regret incurred before identifying the CW, a characteristic unique to this scenario. In the asymptotic regime, the regret scales as $O\left(K\chi(\mathcal{G}_{\gamma})\log T\right)$, consistent with distributed approaches in the multiplayer MAB setting \citep{dubey2020cooperative}. For the no-communication case ($\gamma=0$), this becomes an $O\left(MK\log T\right)$ bound, while for complete communication ($\gamma=D$), where every two players can communicate, the regret scales optimally as $O\left(K\log T\right)$. 

\paragraph{Non-dominant Regret Term}
\label{par: Non-asymptotic term}
In the finite-time regime, the non-dominant term $\hat{C}$ is non-increasing with $\gamma$, and is determined separately for each graph $\mathcal{G}$ and parameter $\gamma$ by minimization over two competing delay-dependent terms. While providing a direct expression for this term is difficult, in a complete communication setting, it scales like $O\left(MKD+\frac{K^2\log K}{\Delta^2_{1\min}}+MK^2\min\Big\{\frac{\log K}{\Delta^2_{1\min}},D\Big\}\right)$, as discussed in Appendix~\ref{sec:Appendix C: Message Passing RUCB Proofs}. Here, we observe a tradeoff between the delay $D$ and the instance complexity $1/\Delta^2_{1\min}$. For challenging instances where $\Delta^2_{1\min}$ is sufficiently small, $\hat{C}$ scales as $O(K^2\log K/\Delta^2_{1\min})$ compared to the $O(MK^2\log K/\Delta^2_{1\min})$ non-leading term for FYL-RUCB. This observation helps explain the better empirical performance observed in Section~\ref{sec: Experiments}. 
\begin{figure*}
    \begin{minipage}{0.33\textwidth}
        \centering
        \includegraphics[width=\linewidth]{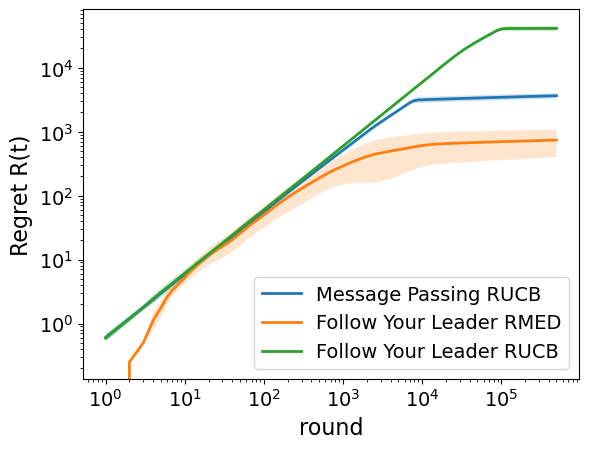}
        \subcaption{Six Rankers}
    \end{minipage}\hfill
    \begin{minipage}{0.33\textwidth}
        \centering
        \includegraphics[width=\linewidth]{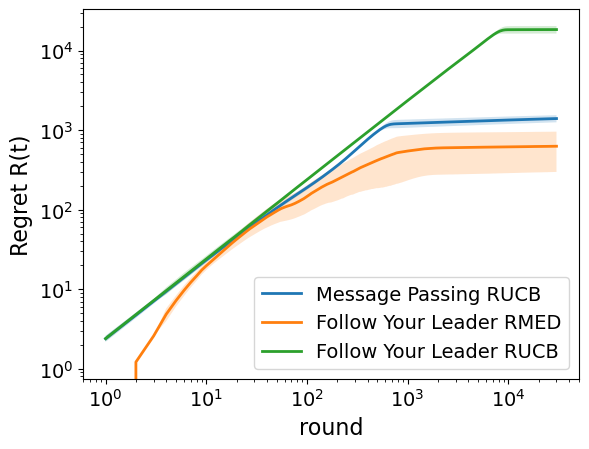}
        \subcaption{Sushi}
    \end{minipage}\hfill
    \begin{minipage}{0.33\textwidth}
        \centering
        \includegraphics[width=\linewidth]{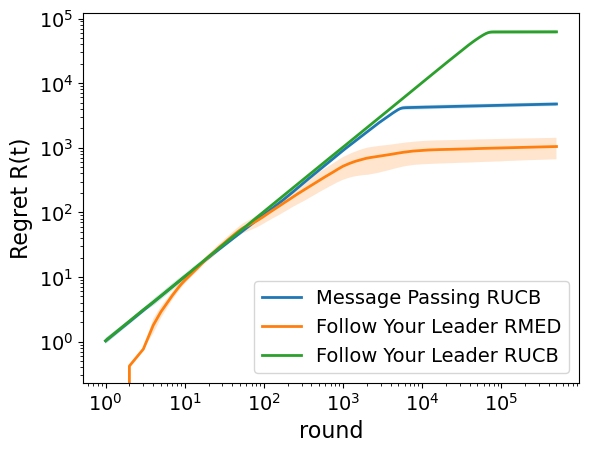}
        \subcaption{Irish}
    \end{minipage}

    \caption{Group regret for all datasets, using a complete communication graph with 10 players.}
    \label{fig: all comparison}
\end{figure*}
\begin{figure*}
    \begin{minipage}{0.33\textwidth}
        \centering
        \includegraphics[width=\linewidth]{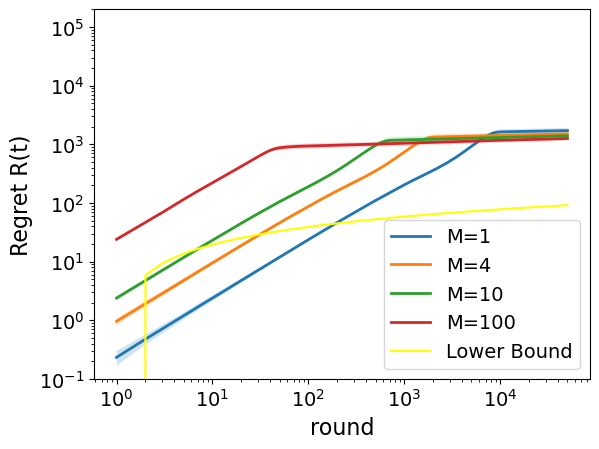}
        \subcaption{Message Passing RUCB}
    \end{minipage}\hfill
    \begin{minipage}{0.33\textwidth}
        \centering
        \includegraphics[width=\linewidth]{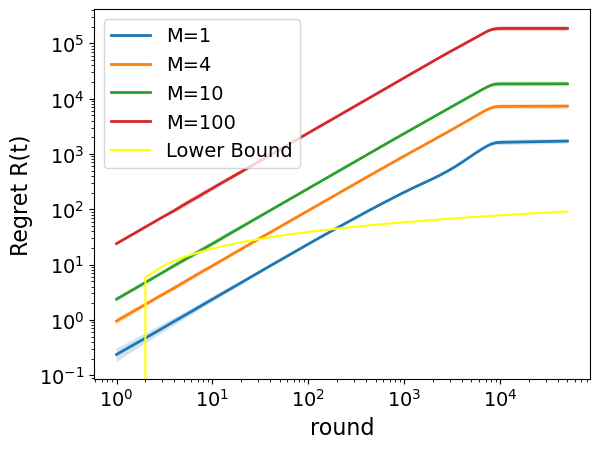}
        \subcaption{Follow Your Leader RUCB}
    \end{minipage}\hfill
    \begin{minipage}{0.33\textwidth}
        \centering
        \includegraphics[width=\linewidth]{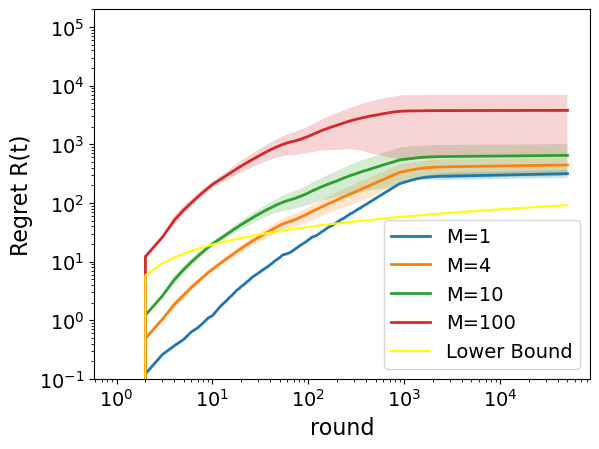}
        \subcaption{Follow Your Leader RMED}
    \end{minipage}\hfill
    \caption{Group regret for the Sushi dataset: a complete graph with a varying number of players.}
    \label{fig: Varying number of players}
\end{figure*}

\paragraph{Proof Outline}
Proving Theorem~\ref{theorem: MP} is not straightforward, since it requires the establishment that each player actively identifies the CW quicker than a single-player counterpart, and that visitation counters within cliques cannot be too large with high probability. To achieve this, the proof relies on two parts. In the first one, we leverage Lemma~\ref{lemma:MP visitation bound} and demonstrate that, owing to CW candidate recommendations, each player identifies the CW within $O\left(\gamma+K^2\gamma'+K^2\log(K)/\Gamma(m,\gamma')\Delta^2_{1\min}\right)$ rounds with high probability for any clique such that $\gamma' \leq \gamma$. In the second part, we decompose the regret into a finite-time term corresponding to $\hat{C}$ and an asymptotic term, and bound them using the arbitrariness of $\gamma'$ and the visitation count bound from Lemma~\ref{lemma:MP visitation bound}. For the asymptotic term, each player consistently selects the CW as the first arm, leading to a high probability regret bound that is later converted into an expected regret bound.

\begin{remark}
The clique analysis in Theorem~\ref{theorem: MP} is implicit, allowing the bound to depend on the optimal clique covering. In contrast, applying a similar clique analysis to Algorithm~\ref{alg:FYLBlacBox}, where each clique has a designated leader, would necessitate actively finding the minimal clique covering, which is an NP-hard problem. Additionally, using several leaders in this setting will result in an a regret that scales with the number of leaders, and not be asymptotically optimal. 

\end{remark}
\section{EXPERIMENTS}
\label{sec: Experiments}
In this section, we present experimental results to assess the empirical performance of Algorithms~\ref{alg:FYLBlacBox} and \ref{alg:MPRUCB}. The simulations utilized three preference matrices:

\textbf{Six rankers:} Used 6 arms, based on the six retrieval functions used in the engine of ArXiv.org \citep{yue2011beat}.\\
\textbf{Sushi:} Derived from the Sushi preference dataset \citep{kamishima2003nantonac}, which contains preference data from 5000 users over 100 types of Sushi. We used a 10-kinds dataset \citep{MaWa13a} and converted it into a preference matrix, where $q_{ij}$ denotes the ratio of users who prefer kind $i$ over $j$.\\
\textbf{Irish election:} Based on the 2002 Dublin Meath election dataset \citep{Jeffery2013}, encompassing 64,081 votes over 14 candidates. We extracted the 10 most preferred candidates and constructed a preference matrix similarly to the Sushi dataset. This dataset was also used in \citet{agarwal2022asymptotically}.

The datasets above all feature a CW. We conducted a comparison between Algorithm~\ref{alg:FYLBlacBox} with both RMED2FH and RUCB as base algorithms, alongside Algorithm~\ref{alg:MPRUCB}. Additionally, we evaluated the performance of the single-player Versatile-DB (VDB) algorithm \citep{saha2022versatile}, which is well-regarded for its strong theoretical guarantees and empirical effectiveness, making it a natural baseline for comparison with our multiplayer algorithms. However, despite its effectiveness, this reduction-to-MAB approach does not track the CW, preventing it from satisfying Assumption~\ref{assumption: BlackBoxAlg}, which is essential for extending it to a FYL multiplayer setting. Moreover, implementing message passing for VDB presents additional challenges, as it requires communicating richer information about loss values, making its adaptation to the multiplayer setting more complex. Additional experimental results and details can be found in Appendix~\ref{sec: Appendix D: More Experiments}.

\paragraph{Algorithmic Comparisons Across Datasets}
\label{subsec:Comparison among algorithms and datasets}
\begin{figure*}[t]
    \begin{minipage}{0.33\textwidth}
        \centering
        \includegraphics[width=\linewidth]{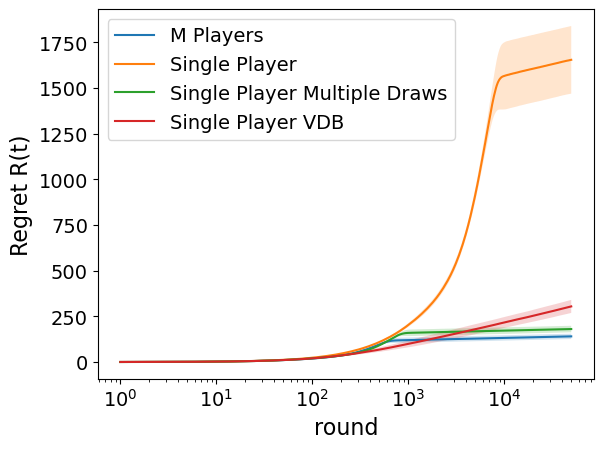}
        \subcaption{Message Passing RUCB}
    \end{minipage}\hfill
    \begin{minipage}{0.33\textwidth}
        \centering
        \includegraphics[width=\linewidth]{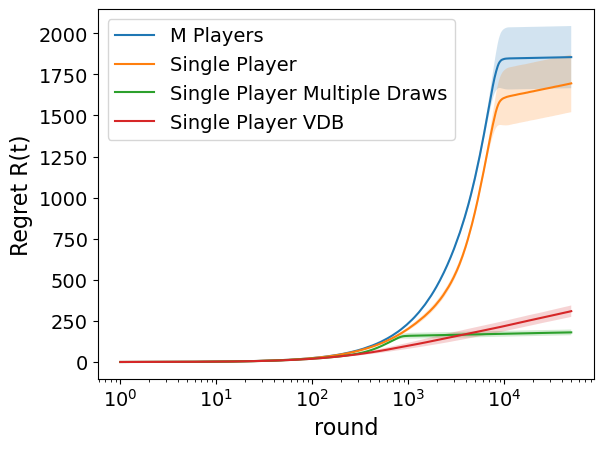}
        \subcaption{Follow Your Leader RUCB}
    \end{minipage}\hfill
    \begin{minipage}{0.33\textwidth}
        \centering
        \includegraphics[width=\linewidth]{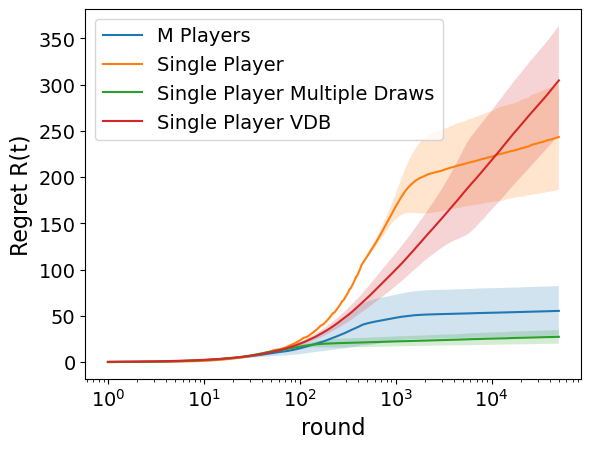}
        \subcaption{Follow Your Leader RMED}
    \end{minipage}

    \caption{Average regret for the Sushi dataset with $M=10$ players: a 10-player group, a single player, a single player with 10 draws in each round, and a single player VDB.}
    \label{fig: Comparison with single player}
\end{figure*}
\begin{figure*}[t]
    \begin{minipage}{0.33\textwidth}
        \centering
        \includegraphics[width=\linewidth]{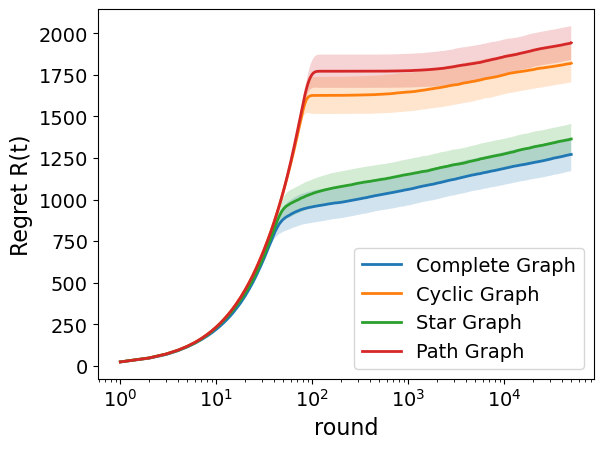}
        \subcaption{Message Passing RUCB}
    \end{minipage}\hfill
    \begin{minipage}{0.33\textwidth}
        \centering
        \includegraphics[width=\linewidth]{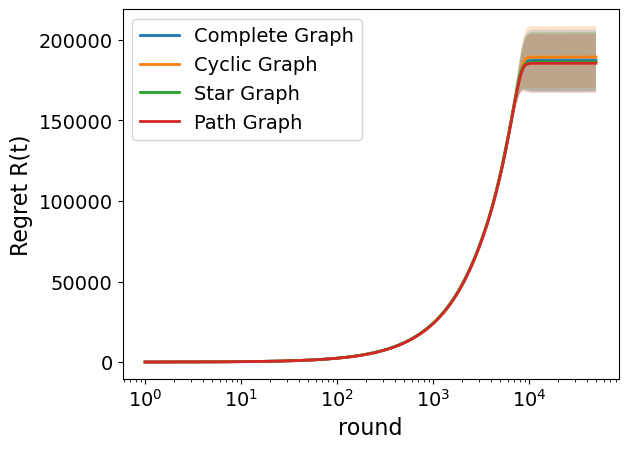}
        \subcaption{Follow Your Leader RUCB}
    \end{minipage}\hfill
    \begin{minipage}{0.33\textwidth}
        \centering
        \includegraphics[width=\linewidth]{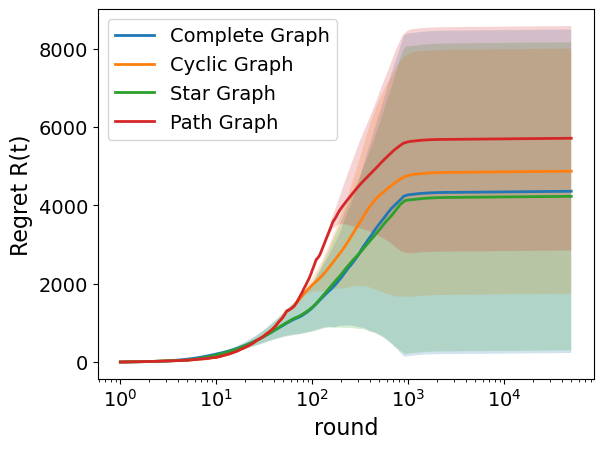}
        \subcaption{Follow Your Leader RMED}
    \end{minipage}\hfill
    \caption{Group regret for the Sushi dataset: different communication graphs with $M=100$ players.}
    \label{fig: Different Graphs}
\end{figure*} 
In Figure~\ref{fig: all comparison} we compare the algorithms across all datasets. The group regret per round is illustrated on a log-log scale, considering a complete communication graph with $M=10$ nodes. For all datasets, we observe both the polynomial finite-time regime, dominated by exploration, and the linear asymptotic one, where the CW has been captured. 
In each dataset, FYLRMED consistently outperforms the other algorithms, while  FYLRUCB is the least favorable. The superior performance of FYLRMED can be attributed to the enhanced exploration-exploitation tradeoff achieved by RMED2FH compared to RUCB, emphasizing that the performance of FYLBB is greatly influenced by the base algorithm. Message-passing RUCB showcases the significance of cooperation during exploration, leading to the early discovery of the CW compared to FYLRUCB, as indicated by the earlier transition into the asymptotic regime. Consequently, the regret is substantially lower, despite both algorithms using a similar decision-making process. 

\paragraph{Experiments with a Varying Number of Players}
\label{subsec:Varying number of players}
In Figure~\ref{fig: Varying number of players}, we depict the group regret using the Sushi dataset on a complete communication graph with $M=1,4,10,100$, along with the lower bound. The results illustrate that, as anticipated by the regret bounds, the regret in the asymptotic regime remains unaffected by $M$. Message-Passing RUCB rapidly identifies the CW as the number of players increases due to collaborative exploration, leading to a regret that slightly reduces with $M$. For the two other algorithms only the leader explores, resulting in late identification of the CW and a regret that increases with the player count.

\paragraph{Comparisons with a Single Player}
\label{subsec:Comparison with single player}

In Figure~\ref{fig: Comparison with single player}, we show the average regret per player in a 10-player group for each algorithm. Additionally, we include the regret of a single player and the regret of a single player capable of making 10 actions in each round, divided by 10. For each figure, we also demonstrate the performance of a single player VDB algorithm with a learning rate of $4/\sqrt{t}$. We utilized the Sushi dataset and used a complete communication graph for the multiplayer setting. Notably, for all algorithms, a single player with only one action per round exhibits inferior performance in the asymptotic regime as implicated by its higher slope. Although less apparent given the time horizon of the experiment, this also holds for Figure~\ref{fig: Comparison with single player}b when $T$ becomes large. Additionally, compared with VDB which incurs a significantly lower regret compared to RUCB and comparable to that of RMED under a single player setting, all multiplayer systems are superior to it both in slope and in the average regret.

These observations underscore the evident advantage of employing multiplayer systems. Conversely, a single player with 10 actions per round simulates a centralized 10-player system, where decision-making by a central entity benefits from all available information. Across all algorithms, the average player in a multiplayer system exhibits a comparable slope in the asymptotic regime to that of the average player in a centralized system, emphasizing their similar performance. This reaffirms that, asymptotically, our algorithms successfully emulate the behavior of a centralized system.

\paragraph{Comparisons Across Different Graph Structures}
\label{subsec:Comparisons across Different Graph Structures}
Figure~\ref{fig: Different Graphs} depicts the group regret per round for the Sushi dataset and all algorithms, employing complete, cycle, star and path communication graphs with 100 players. For FYLBB algorithms, the leader is the central node in the star graph.
Alternatively, it is the outermost node in the path graph. Notably, for Message Passing RUCB a complete graph and a star graph, with diameters of $1$ and $2$ respectively, outperform a cycle graph with a diameter of $50$ and a path graph with a diameter of $99$, due to delayed CW identification. This distinction is less apparent in the case of FYLBB algorithms, where the capture of the CW occurs later on average and dominates the incurred regret. The large variance in Figure \ref{fig: Different Graphs}c is discussed in Appendix~\ref{sec: Appendix D: More Experiments}. 

\section{CONCLUSION AND FUTURE WORK}
\label{sec: Conclusion and  Future Work}
This paper introduced novel approaches to address the multiplayer dueling bandit problem, demonstrating the existence of efficient algorithms despite the increased complexity compared to an MAB setting. We first discussed a versatile black-box algorithm, leveraging either RUCB or RMED as base algorithms to achieve asymptotically optimal regret. However, as demonstrated in the experimental evaluation, finite-time performance can vary depending on the exploration efficiency of the base algorithm. To address this, we devised a novel message-passing protocol with CW recommendations for RUCB, showcasing more consistent performance and quick identification of the CW. All algorithms exhibited asymptotic regret comparable to that of a single player, a result validated theoretically and through simulations.

In future work, one promising avenue is the development of a black-box algorithm capable of compatibility with diverse base algorithms, while also enhancing collaborative exploration. A parallel approach involves devising a multiplayer black-box algorithm using a base multiplayer bandit algorithm, similar to approaches used in single-player dueling bandits \citep{saha2022versatile}. Additionally, making our algorithms more practical for real-world applications could involve incorporating federated learning \citep{wei2023incentivized} and privacy-preserving methods \citep{azize2022privacy}. 

\section*{Acknowledgments}
\label{sec:Acknowledgments}
MS was supported by JST ASPIRE Grant Number JPMJAP2405 and by a grant from Apple, Inc. Any views, opinions, findings, and conclusions or recommendations expressed in this material are those of the authors and should not be interpreted as reflecting the views, policies or position, either expressed or implied, of Apple Inc. JH was supported by JSPS KAKENHI Grant Number JP21K11747.

\bibliography{biblio}
\bibliographystyle{apalike}

\onecolumn
\newpage
\appendix
\section{RELATED WORK}
\label{Appendix: Related Work}
The dueling bandit problem, introduced as a useful variant of the MAB problem in \citet{yue2009interactively}, initially underwent analysis with stringent assumptions about the preference matrix. These assumptions, including stochastic transitivity and stochastic triangle inequality, resulted in regret bounds of $O(K\log T)$ \citep{yue2011beat,yue2012k}. 
Subsequent works relaxed these assumptions by introducing the more realistic CW assumption \citep{zoghi2014relative,komiyama2015regret,chen2017dueling,agarwal2022asymptotically}, which remains widely adopted today. 
Notably, the RUCB algorithm \citep{zoghi2014relative}, based on the UCB approach for MAB, achieves an $O(K\log T)$ regret with high probability. \citet{komiyama2015regret} established a comprehensive lower bound for this setting of the same order and provided the RMED family of algorithms that match it. Recently, \citet{saha2022versatile} demonstrated a general reduction from dueling bandit algorithms to MAB and introduced a best-of-both-worlds algorithm effective for both adversarial and stochastic dueling bandits. Other recent advancements include a batched variant \citep{agarwal2022batched}, a differentially private extension \citep{saha2024dp} and adversarial or non-stationary settings \citep{saha2021adversarial,kolpaczki2022non}. Additionally, in cases where the CW does not exist, alternative definitions of winners, such as the Borda Winner \citep{jamieson2015sparse} and the Copeland Winner \citep{komiyama2016copeland,zoghi2015copeland}, have been considered in dueling bandit literature. Further extensions to the original dueling bandit setting include systems in which more than two arms can be drawn at each round, such as battle of bandits \citep{saha2018battle} and choice bandits \citep{agarwal2020choice}. 

Considerable attention utilizing various approaches has been directed towards cooperative multiplayer MAB scenarios, revealing an $O(K\log T)$ regret lower bound \citep{dubey2020cooperative}. Among them, in an approach called the leader-follower paradigm, players designate one or multiple leaders and base their arm selections on the leaders' instructions. In \citet{kolla2018collaborative} and \citet{landgren2018social}, followers emulate leaders employing a UCB strategy with neighbor-only communication, but the resulting algorithm does not match the lower bound. In the context of a Bernoulli bandit, \citet{wang2020optimal} introduced a distributed parsimonious exploration strategy with message passing akin to ours called DPE2, in which a leader explores using a UCB strategy, and followers exploit a designated arm sent by it. Their approach showcased a regret that aligns with the lower bound, but required a meticulous derivation that cannot be applied in general. Recently, \citet{howson2024quack} suggested a general reduction approach from multiplayer to single player MAB, which can be applied to various settings including multiplayer dueling bandits. However, unlike our work which introduces both a leader-follower approach and a distributed one, their algorithm is only applicable when a leader can be elected, which is not always suitable as discussed in Section~\ref{sec: Followers Your Leader Black Box Algorithm}. In the fully distributed setting, which is more suitable in situations where players are non-synchronized or should have similar performance, both a message-passing approach \citep{dubey2020cooperative,madhushani2021one} and a running-consensus approach \citep{landgren2020distributed} yield a regret that matches the lower bound up to constants. Alternatively, certain approaches adopt a gossiping strategy \citep{chawla2020gossiping, sankararaman2019social}, in which a player can randomly communicate with one other player at each round. 

A different approach to multiplayer MAB assumes that when multiple players select the same arm, a collision occurs, leading to a loss of information \citep{boursier2024survey}. This setting is primarily inspired by the cognitive radio problem, where smart devices must learn the optimal transmission configuration over a shared network \citep{mitola1999cognitive}. In this scenario, information loss arises when multiple users attempt to access the same channel simultaneously. The multiplayer MAB framework with collisions has received significant attention in recent years \citep{boursier2019sic,wang2020optimal,bistritz2018distributed}, with various innovative approaches proposed to address the challenge. In contrast, our work is motivated by the task of fine-tuning large-scale networks with human feedback. This scenario is better captured by the non-collision cooperative setting, where collecting information about the same arm pair does not lead to interference.

Recently, fine-tuning large language models and text-to-image diffusion networks has proven effective and efficient through the use of preference-based human feedback \citep{christiano2017deep,yang2024using}. Approaches include latent reward estimation and direct preference optimization, drawing inspiration from dueling bandit techniques \citep{xia2024beyond,nika2024reward}
\section{COMPUTATIONAL COMPLEXITY, SPACE COMPLEXITY AND COMMUNICATION COST}
\label{Appendix 0A: Computational and Space Complexities}
To efficiently handle message passing, where messages include the original player’s id and the round they were created, players can maintain a table that records the ids of other players and the corresponding time it takes for messages to arrive from each one. This requires no prior knowledge, as the table can be updated when the first message from each player is received. Each player needs $O(M)$ space to store this table, enabling them to disregard redundant messages that may arrive via different paths in the communication graph. As each player can send up to $\gamma$ messages per round (including forwarded messages), the computational complexity of message passing per player is $O(\gamma M)$, due to the need to filter out duplicate messages from neighboring players.

For Algorithm~\ref{alg:FYLBlacBox} initialized with the leader election algorithm from Algorithm~\ref{alg:Simple Leader Election},  the space complexity per round during the leader election phase increases by an additive factor 
$O(M)$ relative to the base algorithm, while the computational complexity grows by $O(M^2)$ due to message passing. Once the leader is elected, it only initiates communication with its neighbors, so the space complexity remains the same as in the base algorithm, and the computational complexity per round increases by an additive factor of $O(M)$. Followers only need to store the currently recommended arm and handle message passing stemming from the leader, resulting in a space complexity of $O(M)$ and a per-round computational complexity of $O(M)$. In Algorithm~\ref{alg:MPRUCB}, players maintain and update optimistic matrices of size $K^2$ in addition to handling message passing. As a result, the space complexity per player is $O(M+K^2)$, and the per-round computational complexity is $O(\gamma M+K^2)$.

In this work, we define the communication cost as the expected number of communication rounds, where a communication round is a round in which some player communicates. Since messages consist solely of integers, this measure aligns with other commonly used metrics, such as the total number of messages \citep{wang2023achieving} and the total number of bits \citep{wang2020optimal}, differing only by a factor of $M\gamma$ and logarithmic terms. For Algorithm~\ref{alg:FYLBlacBox}, the communication cost is bounded by $T_{\mathrm{LE}}+2f(K,T,Q)$, as shown in the proof for Theorem~\ref{theorem:FYLBlackBox}. Therefore, when $f(K,T,Q)=o(T)$, such as when using the base algorithms in Corollary~\ref{corollary:FYLRMED}, the communication cost is sub-logarithmic in $T$, as is often observed in multiplayer MAB \citep{yang2023cooperative} with a leader. By Lemma~\ref{lemma: RMED RUCB CW mistake bound}, the communication cost of FYL-RUCB is independent of the horizon, while for FYL-RMED2FH it is $O(\log\log T)$. Similar to other distributed multiplayer MAB algorithms based on UCB \citep{landgren2020distributed},  the communication cost for Algorithm~\ref{alg:MPRUCB} is linear with $T$. Nevertheless, the computation cost per round remains low, as each message only contains a fixed number of integers. In comparison, running consensus algorithms \citep{landgren2020distributed} in which each player shares the entire preference estimation matrix incurs a computation cost of $O(MK^2)$. Additionally, these algorithms either need to send non-integer numbers in the form of the preference matrix, or matrices with integers of the order $O(T)$ in the form of total visitation counters and arm comparison counters.  
\section{LOWER BOUND PROOF}
\label{Appendix A: Lower Bound Proof}
This section consists of two subsections. In Section~\ref{subsec: A Canonical Probability Space}, we establish a precise definition for a canonical probability space tailored to the multiplayer dueling bandit scenario. Following this foundational definition, Section~\ref{subsec: proof for LB theorem} presents a comprehensive proof for the lower bound encapsulated in Theorem~\ref{theo:DB Multi-Player Lower Bound}. To ensure clarity and coherence, we revisit key definitions introduced in Sections~\ref{sec: Problem Formulation} and \ref{sec:Lower Bound}.
\subsection{A Canonical Probability Space}
\label{subsec: A Canonical Probability Space}
\begin{definition}
	\label{def:Dueling Bandit Environment}
	A stochastic dueling bandit environment is a collection of distributions $\nu=\left(P_{ij}:i,j \in [K]\right)$, where $K$ is the number of available actions, and $P_{ij}$  is a Bernoulli distribution with mean $q_{ij}$, such that $q_{ij}=1-q_{ji}$. 
\end{definition}
In this context, the preference matrix $Q=(q_{ij})\in \mathbb{R}^{K\times K}$ represents a given environment. Next, we define a class of preference matrices.
\begin{definition}
	\label{def:Dueling Bandit Environment Classes}
	A dueling bandit environment class is $\mathcal{Q}=(Q)$ is some collection of preference matrices. In particular, we define,
	\begin{enumerate}[label=(\alph*)]
	    \item The class of preference matrices with total ordering: $ \mathcal{Q}_\mathrm{o}=\{Q~|~i \prec j \Leftrightarrow q_{ij}<1/2\}$.
	   \item The class of preference matrices with a Condorcet Winner (CW): $\mathcal{Q}_{\mathrm{CW}}=\{Q~|~\exists i:q_{ij}>1/2 ~\forall j \neq i\}$.
	\end{enumerate}
\end{definition}
Note that $\mathcal{Q}_{\mathrm{o}} \subset \mathcal{Q}_{\mathrm{CW}}$. Now, let $T$ be the time horizon. We consider $M$ players communicating via a connected, undirected graph $\mathcal{G}=(\mathcal{V},\mathcal{E})$. Communication is bidirectional, and any message sent from player $m$ may be obtained by player $m'$ after $d(m,m')-1$ rounds of the bandit problem, where $d(m,m')$ denotes the length of the shortest path between players $m$ and $m'$ on $\mathcal{G}$. Let $(i_m(t),j_m(t))$ denote the pair drawn by player $m$ at time $t$, and
$r_{m}(t)$ denote the corresponding reward. The power graph of order $\gamma$ of $\mathcal{G}$, denoted by $\mathcal{G}_{\gamma}$, is defined to include an edge $(m,m')$ if there exists a path of length at most $\gamma$ in $\mathcal{G}$ between players $m$ and $m'$. the neighborhood of $m$ in $\mathcal{G}_{\gamma}$ is given by $\mathcal{N}_{\gamma}(m)$, including player $m$ itself. We establish a communication protocol as follows.
\begin{assumption}
	\label{ass:Communication Protocol DB}
	Players can communicate in the following manner:
	\begin{itemize}
	    \item Any player $m$ is capable of sending a message $x_{m}^{m'}(t)$ to any other player $m'\in \mathcal{N}_{\gamma}(m)$ at time $t$, and this message may be received by player $m'$ at time $t+\min(0,d(m,m')-1)$. This is done via a message-passing protocol as described in Section~\ref{sec: Problem Formulation}
	    \item The message $x_{m}^{m'}(t)$ is a function of the arms-reward triplets of player $m$ up to and including time $t$, i.e. $x_{m}^{m'}(t)=F_{m,t}^{m'}\left(i_m(1),j_m(1),r_m(1),\ldots,i_m(t),j_m(t),r_m(t)\right)$ for any deterministic Borel function $F_{m,t}^{m'}:\mathbb{R}^{3t}\rightarrow\mathbb{R}^L$, with $L$ being a nonnegative integer. 
	\end{itemize}
\end{assumption}
To define a measurable probability space for this setting, let us treat the players as ordered, such that player $1$ is the first one and player $M$ is the last. We define the outcome until time $t$ and player $m$ as the ordered set $H_{m,t}=(i_1(1),j_1(1),r_1(1),\ldots,i_m(t),j_m(t),r_m(t))$, the outcome space as $\Omega_{m,t}=([K]^{2}\times \mathbb{R})^{M(t-1)+mt}$, and the corresponding sigma-algebra as $\mathcal{F}_{m,t}=\mathcal{B}(\Omega_{m,t})$. Here, $\mathcal{B}(\cdot)$ stands for the Borel sigma-algebra. Using these, we define a measurable space for the entire multiplayer dueling bandit problem as $\left(\Omega_{M,T},\mathcal{F}_{M,T}\right)$, with the random variables of the drawn arms and obtained rewards as
\begin{equation*}
    \begin{aligned}
    i_m(t)\left(i_{1,1},j_{1,1},x_{1,1},\ldots,i_{M,T},j_{M,T},r_{M,T}\right)&=i_{m,t}~,\\
    j_m(t)\left(i_{1,1},j_{1,1},x_{1,1},\ldots,i_{M,T},j_{M,T},r_{M,T}\right)&=j_{m,t}~,\\
    r_m(t)\left(i_{1,1},j_{1,1},x_{1,1},\ldots,i_{M,T},j_{M,T},r_{M,T}\right)&=r_{m,t}.
    \end{aligned}
\end{equation*}
For clerity, we also define the player-specific measurable space as $\left(\Omega_{T}^{m},\mathcal{F}_{T}^{m}\right)$, where $H_{t}^{m}=(i_m(1),j_m(1),r_m(1),\ldots,i_m(t),j_m(t),r_m(t))$, $\Omega_{t}^{m}=\left([K]^{2}\times \mathbb{R}\right)^{t}$, and $\mathcal{F}_{t}^{m}=\mathcal{B}(\Omega_{t}^{m})$.
Next, we define the multiplayer dueling bandit policy, which can be separated into a communication policy and an action policy. This separation is motivated by the fact that both aspects are under the control of the learner.
\begin{definition}
	\label{def:DB Multi-Player Communication Policy}
	A communication policy $\tilde{\Pi}_{\mathrm{comm},\gamma}$ for the multiplayer dueling bandit setup is the sequence of messages $(F_{m,t}^{m'})_{m,m',t}$ from Assumption \ref{ass:Communication Protocol DB}, where $m \in [M],t \in [T]$ and $m'\in \mathcal{N}_{\gamma}(m) ~\forall m\in[M]$.
\end{definition}
\begin{definition}
	\label{def:DB Multi-Player Action Policy}
	An action policy $\tilde{\Pi}$ for the multiplayer dueling bandit setup is a sequence $(\tilde{\Pi}_{t})_{t=1}^{T}$ where $\tilde{\Pi}_{t}=(\tilde{\Pi}_{t}^{m})$, and $\tilde{\Pi}_{t}^{m}$ is a probability kernel from $\left(\Omega_{t-1}^{m}\times \mathbb{R}^{L_{m,t}^{\gamma}},\mathcal{B}\left(\Omega_{t-1}^{m}\times \mathbb{R}^{L_{m,t}^{\gamma}}\right)\right)$ to $([K]^{2},2^{[K]^{2}})$, where $L_{m,t}^{\gamma}=L\sum_{m'\in \mathcal{N}_{\gamma}(m)\backslash \{m\}}(t-d(m,m'))$. We define $\tilde{\pi}_{t}=(\tilde{\pi}_{t}^{m})$ as the corresponding density with respect to the counting measure on $([K]^{2},2^{[K]^{2}})$, so that $\forall i,j \in [K]$,
	\begin{equation*}
	    \tilde{\pi}_{t}^{m}\left(i,j|h_{t-1}^{m},z_{t}^{m}\right)=\tilde{\Pi}_{t}^{m}\left(\{i,j\}|h_{t-1}^{m},z_{t}^{m}\right).
	\end{equation*}
	Here, where $z_{t}^{m}$ stands for all the messages received by player $m$ until round $t$. 
\end{definition}
The communication policy $\tilde{\Pi}_{\mathrm{comm},\gamma}$ and the action policy $\tilde{\Pi}$ as defined above illustrate the nature through which a learner operates, are easy to extract from an algorithm, and will help define the probability density function of the outcome. However, we can combine them into a more convenient policy definition that only takes into account the outcomes in $\Omega_{M,T}$.
\begin{definition}
	\label{def:DB Multi-Player Policy}
	A policy $\Pi$ for the multiplayer dueling bandit setup is a sequence $(\Pi_{t})_{t=1}^{T}$ where $\Pi_{t}=(\Pi_{t}^{m})$, and $\Pi_{t}^{m}$ is a probability kernel from $\left(\Omega_{t-1}^{m}\times \prod_{m'\in \mathcal{N}_{\gamma}(m)\backslash \{m\}}\Omega_{t-d(m,m')}^{m'},\mathcal{B}\left(\Omega_{t-1}^{m}\times\prod_{m'\in \mathcal{N}_{\gamma}(m)\backslash \{m\}}\Omega_{t-d(m,m')}^{m'}\right)\right)$ to $([K]^{2},2^{[K]^{2}})$. We define $\pi_t=(\pi_{t}^{m})$ as the corresponding density with respect to the counting measure on $([K]^{2},2^{[K]^{2}})$, so that$\forall i,j \in [K]$
	\begin{equation*}
	    \pi_{t}^{m}\left(i,j\Bigg|h_{t-1}^{m}\times\prod_{m'\in \mathcal{N}_{\gamma}(m)\backslash \{m\}}h_{t-d(m,m')}^{m'}\right)=\tilde{\pi}_{t}^{m}\left(i,j|h_{t-1}^{m},z_{t}^{m}\right).
	\end{equation*}
\end{definition}
We are now ready to define the probability measure on $\left(\Omega_{M,T},\mathcal{F}_{M,T}\right)$, which can be done in two ways similar to the MAB setting \citep{lattimore2020bandit}. As a first definition, we require that - 
\begin{itemize}
    \item The conditional distribution of $i_m(t),j_m(t)$ given $i_1(1),\ldots,r_{m-1}(t)$ (in case $m> 1$, or $i_1(1),\ldots,r_M(t-1)$ otherwise) is $\Pi_{t}^{m}$ almost surely.
    \item The conditional distribution of $r_m(t)$ given $i_1(1),\ldots,r_{m-1}(t),i_m(t),j_m(t)$ is $P_{i_m(t),j_m(t)}(r_m(t))$ almost surely.
\end{itemize}
This defines a unique probability measure on $\left(\Omega_{M,T},\mathcal{F}_{M,T}\right)$. Alternatively, as a second definition, we can also define a probability measure with respect to a probability density function (pdf). Let $\lambda$ be a $\sigma$-finite measure on $(\mathbb{R},\mathcal{B}(\mathbb{R}))$ for which $P_{ij}$ is absolutely continuous with respect to $\lambda$ for all $i,j\in [K]$. denote $p_{ij}=dP_{ij}/d\lambda$, and let $\rho$ be the counting measure on $([K]^{2},2^{[K]^{2}})$ (since $P_{ij}$ is Bernoulli, we can take $\lambda$ to be the Lebesgue measure, so that $p_{ij}(x)=q_{ij}x+(1-q_{ij})x$). Utilizing Fubini's theorem and the properties of the Radon-Nikodym derivative, the pdf of an outcome is,
\begin{equation*}
\begin{aligned}
   &p_{\nu \pi}(i_{1,1},j_{1,1},r_{1,1},\ldots,i_{M,T},j_{M,T},r_{M,T})\\
   &=\prod_{t=1}^{T}\prod_{m=1}^{M}\pi_{t}^{m}\left(i_{m,t},j_{m,t}|h_{t-1}^{m}\times\prod_{m'\in \mathcal{N}_{\gamma}(m)\backslash \{m\}}h_{t-d(m,m')}^{m'}\right)p_{i_{m,t},j_{m,t}}(r_{m,t}),
\end{aligned}
\end{equation*}
and the measure can be calculated as,
\begin{equation*}
\begin{aligned}
    \mathbb{P}_{\nu\pi}(B)=&\int_{B}\left(p_{\nu\pi}(i_{1,1},j_{1,1},r_{1,1},\ldots,i_{M,T},j_{M,T},r_{M,T})(\rho \times \lambda)^{MT} \right.\\
    &\quad\left. \times d(i_{1,1},j_{1,1},r_{1,1},\ldots,i_{M,T},j_{M,T},r_{M,T})\right)~\forall B\in \mathcal{F}_{M,T}.
\end{aligned}
\end{equation*}
As mentioned before, these two definitions result in the same probability measure.

\subsection{Proof for Theorem~\ref{theo:DB Multi-Player Lower Bound}}
\label{subsec: proof for LB theorem}
First, we present a useful decomposition for the group regret within the context of multiplayer dueling bandits. Throughout this section, we will frequently employ the notation $\mathbb{E}\mathcal{R}_{\mathcal{G},T}(\pi,\nu)$ instead of $\mathbb{E}\mathcal{R}(T)$ to underscore its dependence on various terms.
\begin{equation*}
\begin{aligned}
   \mathbb{E}\mathcal{R}_{\mathcal{G},T}(\pi,\nu)&=\mathbb{E}_{\nu}\left[\sum_{t=1}^{T}\sum_{m=1}^{M}\frac{\Delta_{1i_m(t)}+\Delta_{1j_m(t)}}{2}\right]\\
   &=\frac{1}{2}\sum_{i\in [K]}\sum_{j \in [K]\setminus\{i\}}\frac{\Delta_{1i}+\Delta_{1j}}{2}\mathbb{E}_{\nu}N_{ij}(T)+\sum_{i\in [K]}\Delta_{1i}\mathbb{E}_{\nu}N_{ii}(T).
\end{aligned}
\end{equation*}
Here, the subscript $\nu$ signifies that the expectation is evaluated for bandit environment $\nu$. In adittion, $N_{ij}(T)=\sum_{t=1}^{T}\sum_{m=1}^{M}\left(\mathds{1}\{i_m(t)=i,j_m(t)=j\}+\mathds{1}\{i_m(t)=j,j_m(t)=i\}\right)$ for $i \neq j$ and $N_{ii}(T)=\sum_{t=1}^{T}\sum_{m=1}^{M}\mathds{1}\{i_m(t)=i,j_m(t)=i\}\}$ represent the total number of visits to arms $(i,j)$ by time $T$ without order. We proceed to define the class of consistent policies, a class for which the lower bound holds.
\begin{definition}
	\label{def:DB Multi-Player Consistent Policy}
	A policy $\pi$ for the multiplayer dueling bandit setup is called consistent over a class of bandits $\mathcal{Q}$ if for all $Q \in \mathcal{Q}$ and $p>0$, it holds that,
	\begin{equation*}
	    \lim_{T \rightarrow \infty}\frac{\mathbb{E}\mathcal{R}_{\mathcal{G},T}(\pi,\nu)}{T^{p}}=0.
	\end{equation*}
	The class of consistent policies over $\mathcal{Q}$ is denoted by $\Pi_{\mathrm{cons}}(\mathcal{Q})$.
\end{definition}
Next, we present a divergence decomposition lemma tailored to the multiplayer dueling bandit setting. This will be used to prove the lower bound. 
\begin{lemma}
	\label{lemma:DB Multi-Player Divergence Decomposition}
	Let $\nu=(P_{ij})$ and $\nu'=(P_{ij}^{'})$ be two stochastic dueling bandit environments. Fix some multiplayer policy $\pi$ such that Assumption \ref{ass:Communication Protocol DB} holds, and let $\mathbb{P}_{\nu\pi}$ and $\mathbb{P}_{\nu'\pi}$ denote the probability measures for the canonical bandit models induced by the $T$-round interconnection of $\pi$ and $\nu,\nu'$ respectively. Then, 
	\begin{equation*}    \mathrm{KL}\left(\mathbb{P}_{\nu\pi},\mathbb{P}_{\nu'\pi}\right)=\frac{1}{2}\sum_{i\in [K]}\sum_{j\in [K]\setminus \{i\}}\mathbb{E}_{\nu}[N_{ij}(T)]\mathrm{KL}(q_{ij},q_{ij}^{'}), 
	\end{equation*}
	where $\mathrm{KL}(\mathbb{P},\mathbb{Q})$ stands for the KL divergence between probability measures $\mathbb{P},\mathbb{Q}$
	\begin{equation*}
	    \mathrm{KL}\left(\mathbb{P},\mathbb{Q}\right)=\mathbb{E}_{\mathbb{P}}\left[\log\left(\frac{d\mathbb{P}}{d\mathbb{Q}}\right)\right], 
	\end{equation*}
	and $\mathrm{KL}(p,q)$ stands for the KL divergence between two Bernoulli measures with means $p,q$ respectively.
\end{lemma}
\begin{proof}
Assume that $\mathrm{KL}\left(q_{ij},q_{ij}^{'}\right)<\infty$ for all $i,j \in [K]$. It follows from the KL-divergence definition that $P_{ij}\ll P_{ij}^{'}$. Next, define the measure $\lambda=\sum_{i=1}^{K}(P_{ij}+P_{ij}^{'})$ for which $P_{ij},P_{ij}^{'}\ll \lambda$ for all $i,j \in [K]$. 

It is evident that the policy terms $\pi_{t}^{m}$ remain consistent across bandit environments $\nu$ and bandit $\nu'$ when our interest lies in the same outcome. This consistency arises from Assumption \ref{ass:Communication Protocol DB}, which ensures that messages are deterministic functions of the outcome and are otherwise independent of the environment. Additionally, a policy is uniquely determined by action-reward pairs and messages. Therefore, using the pdf of the outcome with respect to the product measure $(\rho \times \lambda)^{MT}$ for each environment,

\begin{equation*}
    \log\frac{d\mathbb{P}_{\nu\pi}}{d\mathbb{P}_{\nu'\pi}}(i_{1,1},\ldots,r_{M,T})=\sum_{t=1}^{T}\sum_{m=1}^{M}\log\frac{p_{i_{m,t},j_{m,t}}(r_{m,t})}{p_{i_{m,t},j_{m,t}}^{'}(r_{m,t})}~,
\end{equation*}
and
\begin{equation*}
    \mathbb{E}_{\nu}\left[\log\frac{d\mathbb{P}_{\nu\pi}}{d\mathbb{P}_{\nu'\pi}}(i_{1,1},\ldots,r_{M,T})\right]=\sum_{t=1}^{T}\sum_{m=1}^{M}\mathbb{E}_{\nu}\left[\log\frac{p_{i_{m,t},j_{m,t}}(r_{m,t})}{p_{i_{m,t},j_{m,t}}^{'}(r_{m,t})}\right].
\end{equation*}
For every term in the sum we have that,
\begin{equation*}
\begin{aligned}
    \mathbb{E}_{\nu}\left[\log\frac{p_{i_{m,t},j_{m,t}}(r_{m,t})}{p_{i_{m,t},j_{m,t}}^{'}(r_{m,t})}\right]&=\mathbb{E}_{\nu}\left[\mathbb{E}_{\nu}\left[\log\frac{p_{i_{m,t},j_{m,t}}(r_{m,t})}{p_{i_{m,t},j_{m,t}}^{'}(r_{m,t})}\Big{|}i_{m,t},j_{m,t}\right]\right]\\
    &=\mathbb{E}_{\nu}\left[\mathrm{KL}(P_{i_{m,t},j_{m,t}},P_{i_{m,t},j_{m,t}}^{'})\right].
\end{aligned}
\end{equation*}
Returning to the previous equation,
\begin{equation*}
    \begin{aligned}
    \mathbb{E}_{\nu}&\left[\log\frac{d\mathbb{P}_{\nu\pi}}{d\mathbb{P}_{\nu'\pi}}(i_{1,1},\ldots,r_{M,T})\right]=\sum_{t=1}^{T}\sum_{m=1}^{M}\mathbb{E}_{\nu}\left[\mathrm{KL}(P_{i_{m,t},j_{m,t}},P_{i_{m,t},j_{m,t}}^{'})\right]\\
    &=\sum_{i,j=1}^{K}\mathbb{E}_{\nu}\left[\sum_{t=1}^{T}\sum_{m=1}^{M}\mathds{1}\{(i_{m,t}=i,j_{m,t}=j)\}\mathrm{KL}(P_{ij},P_{ij}^{'})\right]\\
    &=\frac{1}{2}\sum_{i\in [K]}\sum_{j\in [K]\setminus \{i\}}\mathbb{E}_{\nu}[N_{ij}(T)]\mathrm{KL}(P_{ij},P_{ij}^{'})+\sum_{i\in [K]}\mathbb{E}_{\nu}[N_{ii}(T)]\mathrm{KL}(P_{ij},P_{ij}^{'})
    \\
    &=\frac{1}{2}\sum_{i\in [K]}\sum_{j\in [K]\setminus \{i\}}\mathbb{E}_{\nu}[N_{ij}(T)]\mathrm{KL}(q_{ij},q_{ij}^{'})+\sum_{i\in [K]}\mathbb{E}_{\nu}[N_{ii}(T)]\mathrm{KL}(q_{ii},q_{ii}^{'})\\
    &=\frac{1}{2}\sum_{i\in [K]}\sum_{j\in [K]\setminus \{i\}}\mathbb{E}_{\nu}[N_{ij}(T)]\mathrm{KL}(q_{ij},q_{ij}^{'}),
    \end{aligned}
\end{equation*}
where in the last transition we used the fact that $q_{ii}=q_{ii}^{'}=1/2$. Given that  $d\mathbb{P}_{\nu\pi}/d\mathbb{P}_{\nu'\pi} < \infty$, it holds that,
\begin{equation*}
    \mathrm{KL}\left(\mathbb{P}_{\nu\pi},\mathbb{P}_{\nu'\pi}\right)=\mathbb{E}_{\nu}\left[\log\frac{d\mathbb{P}_{\nu\pi}}{d\mathbb{P}_{\nu'\pi}}\right].
\end{equation*}
It is evident that in our case the expectation is finite, so 
\begin{equation*}
\mathrm{KL}\left(\mathbb{P}_{\nu\pi},\mathbb{P}_{\nu'\pi}\right)=\frac{1}{2}\sum_{i\in [K]}\sum_{j\in [K]\setminus \{i\}}\mathbb{E}_{\nu}[N_{ij}(T)]\mathrm{KL}(q_{ij},q_{ij}^{'}).
\end{equation*}
In case $\mathrm{KL}(q_{ij},q_{ij}^{'})=\infty$ for some $i,j\in[K]$, this relation still holds since both sides of the equation become infinite. 
\end{proof}
We also use the following lemma.
\begin{lemma}
\label{lemma:DB Multi-Player Visitations Lower Bound}
For any consistent algorithm on $\mathcal{Q}_{\mathrm{CW}}$ and $Q \in \mathcal{Q}_{\mathrm{CW}}$, and for any arm $i \in [K]\setminus \{1\}$ the following holds.
\begin{equation*}
        \liminf_{T \rightarrow \infty}\frac{\mathbb{E}\left[\sum_{j \in \mathcal{O}_{i}}\mathrm{KL}(q_{ij},1/2)N_{ij}(T)\right]}{\log T}\geq 1.
    \end{equation*}
\end{lemma}
\begin{proof}
    Fix some arm $i \in [K]\setminus \{1\}$ and define $\mathcal{O}_{i}^{'}=\{j~|~j\in[K],q_{ij}\leq 1/2\}$. We consider a challenging instance $Q^{'}$, which is defined such that the means $q_{ij}^{'}$ for $j\in \mathcal{O}_{i}^{'}$ satisfy $\mathrm{KL}(q_{ij},q_{ij}^{'})=\mathrm{KL}(q_{ij},1/2)+\epsilon$ with $q_{ij}^{'}>1/2$, for some $\epsilon>0$ (we leave $q_{ii}^{'}=q_{ii}=1/2$). The rest of the means remain the same as $q_{ij}$. For this new instance, $Q^{'} \in \mathcal{Q}_{\mathrm{CW}}$ and arm $i$ is the CW. 

    It is important to highlight that a proof scheme akin to ours that is used for MAB \citep{lattimore2020bandit} is applicable only to an unstructured class of bandits. This is due to the necessity of the unstructured assumption to construct the challenging instance $\nu^{'}$. The unstructured assumption becomes crucial when attempting to change the mean of a single arm, as the presence of interdependence among arms restricts such modifications. In the dueling bandit scenario, we can directly provide a challenging instance within the structured class, even though the arms' means may be correlated. This distinction allows for a more flexible approach to constructing challenging instances within the dueling bandit framework.

    Note that since only the means $q_{ij}$ for $j \in \mathcal{O}_{i}^{'}$ are different in the two environments, Lemma~\ref{lemma:DB Multi-Player Divergence Decomposition} states that,
    \begin{equation*}
    \mathrm{KL}\left(\mathbb{P}_{\nu\pi},\mathbb{P}_{\nu'\pi}\right)= \sum_{j \in \mathcal{O}_{i}^{'}}\mathrm{KL}(q_{ij},q_{ij}^{'})\mathbb{E}_{\nu}\left[N_{ij}(T)\right]   \leq \sum_{j \in \mathcal{O}_{i}^{'}}(\mathrm{KL}(q_{ij},1/2)+\epsilon)\mathbb{E}_{\nu}\left[N_{ij}(T)\right].
    \end{equation*}
    From the Bretagnolle-Huber inequality, for any event $A\in\mathcal{F}_{M,T}$,
    \begin{equation*}
    \begin{aligned}
        \mathbb{P}_{\nu\pi}(A)+\mathbb{P}_{\nu'\pi}(A^{c})&\geq \frac{1}{2}\exp\left(-\mathrm{KL}\left(\mathbb{P}_{\nu\pi},\mathbb{P}_{\nu'\pi}\right)\right)\\
        &\geq \frac{1}{2}\exp\left(-\sum_{j \in \mathcal{O}_{i}^{'}}(\mathrm{KL}(q_{ij},1/2)+\epsilon)\mathbb{E}_{\nu}\left[N_{ij}(T)\right]\right).
    \end{aligned}
    \end{equation*}
    Choose $A=\{N_{ii}(T)>MT/2\}$, and let $\mathcal{R}_{\mathcal{G}}:=\mathcal{R}_{\mathcal{G},T}(\pi,\nu),\mathcal{R}_{\mathcal{G}}^{'}:=\mathcal{R}_{\mathcal{G},T}(\pi,\nu')$. By the Markov inequality,
    \begin{equation*}
    \mathcal{R}_{\mathcal{G}}\geq \Delta_{1i}\mathbb{E}_{\nu}\left[N_{ii}(T)\right] \geq \frac{MT\Delta_{1i}}{2}\mathbb{P}_{\nu\pi}\left(N_{ii}(T) \geq \frac{MT}{2}\right)=\frac{MT\Delta_{1i}}{2}\mathbb{P}_{\nu\pi}(A).
    \end{equation*}
    Additionally, the regret for the challenging instance is
    \begin{equation*}
    \begin{aligned}
       \mathcal{R}_{\mathcal{G}}^{'}&=\frac{1}{2}\sum_{l\in [K] \setminus \{i\}}\sum_{j \in [K]\setminus\{i,l\}}\frac{\Delta_{il}^{'}+\Delta_{ij}^{'}}{2}\mathbb{E}_{\nu^{'}}N_{lj}(T)\\
       &+\sum_{j \in [K]}\Delta_{ij}^{'}\mathbb{E}_{\nu^{'}}N_{jj}(T)+\sum_{j\in [K]\setminus \{i\}}\frac{\Delta_{ij}^{'}}{2}\mathbb{E}_{\nu^{'}}N_{ij}(T).
    \end{aligned}
    \end{equation*}
    By definition, since arm $i$ is the CW for the challenging instance, the minimal gap with respect to it $\Delta_{min}^{'}=\min_{j\in [K] \setminus \{i\}}\Delta_{ij}^{'}$ is positive, so
    \begin{equation*}
    \begin{aligned}
       \mathcal{R}_{\mathcal{G}}^{'}&\geq\frac{1}{2}\sum_{l\in [K] \setminus \{i\}}\sum_{j \in [K]\setminus\{i,l\}}\frac{\Delta_{min}^{'}/2+\Delta_{min}^{'}/2}{2}\mathbb{E}_{\nu^{'}}N_{lj}(T)\\
       &+\sum_{j \in [K]\setminus \{i\}}(\Delta_{min}^{'}/2)\mathbb{E}_{\nu^{'}}N_{jj}(T)+\sum_{j\in [K]\setminus \{i\}}\frac{\Delta_{min}^{'}}{2}\mathbb{E}_{\nu^{'}}N_{ij}(T)\\
       &=\frac{\Delta_{min}^{'}}{2}\mathbb{E}_{\nu^{'}}\left[\frac{1}{2}\sum_{l\in [K] \setminus \{i\}}\sum_{j \in [K]\setminus\{i,l\}}N_{lj}(T)+\sum_{j \in [K]\setminus \{i\}}N_{jj}(T)+\sum_{j\in [K]\setminus \{i\}}N_{ij}(T)\right]\\
       &=\frac{\Delta_{min}^{'}}{2}\mathbb{E}_{\nu^{'}}\left[MT-N_{ii}(T)\right] \geq \frac{\Delta_{min}^{'}}{2} (MT-\frac{MT}{2})\mathbb{P}_{\nu^{'}\pi}\left(MT-N_{ii}(T) \geq MT-\frac{MT}{2}\right)\\
       &=\frac{MT\Delta_{min}^{'}}{4}\mathbb{P}_{\nu^{'}\pi}(A^{c}).
    \end{aligned}
    \end{equation*}
    Putting it all together, 
    \begin{equation*}
    \begin{aligned}
          \mathcal{R}_{\mathcal{G}}+ \mathcal{R}_{\mathcal{G}}^{'} &\geq 
          \frac{MT\Delta_{1i}}{2}\mathbb{P}_{\nu\pi}(A)+\frac{MT\Delta_{min}^{'}}{4}\mathbb{P}_{\nu^{'}\pi}(A^{c})  \\
          &\geq \frac{MT}{4}\min\{2\Delta_{1i},\Delta_{min}^{'}\}\left(\mathbb{P}_{\nu\pi}(A)+\mathbb{P}_{\nu'\pi}(A^{c})\right)\\
          &\geq \frac{MT}{8}\min\{2\Delta_{1i},\Delta_{min}^{'}\}\exp\left(-\sum_{j \in \mathcal{O}_{i}^{'}}(\mathrm{KL}(q_{ij},1/2)+\epsilon)\mathbb{E}_{\nu}\left[N_{ij}(T)\right]\right).
    \end{aligned}
     \end{equation*}
    Rearranging the previous inequality,
    \begin{equation*}
    \sum_{j \in \mathcal{O}_{i}^{'}}(\mathrm{KL}(q_{ij},1/2)+\epsilon)\mathbb{E}_{\nu}\left[N_{ij}(T)\right]\geq \log \left(\frac{MT\min\{2\Delta_{1i},\Delta_{min}^{'}\}}{8\left(\mathcal{R}_{\mathcal{G}}+ \mathcal{R}_{\mathcal{G}}^{'}\right)}\right),
    \end{equation*}
    and:
    \begin{equation*}
    \begin{aligned}
       \liminf_{T\rightarrow \infty} \frac{\sum_{j \in \mathcal{O}_{i}^{'}}(\mathrm{KL}(q_{ij},1/2)+\epsilon)\mathbb{E}_{\nu}\left[N_{ij}(T)\right]}{\log T}&\geq  \liminf_{T\rightarrow \infty} \frac{\log \left(\frac{MT\min\{2\Delta_{1i},\Delta_{min}^{'}\}}{8\left(\mathcal{R}_{\mathcal{G}}+ \mathcal{R}_{\mathcal{G}}^{'}\right)}\right)}{\log T} \\
       &=\left(1-\limsup_{T\rightarrow \infty}\frac{\log \left(\mathcal{R}_{\mathcal{G}}+ \mathcal{R}_{\mathcal{G}}^{'}\right)}{\log T}\right)=1.
    \end{aligned}
    \end{equation*}
    In the last transition we used that fact that $\pi \in \Pi_{\mathrm{cons}}\left(\mathcal{Q}\right)$, so for every $p>0$ there exists some constant $C_p$ such that $\mathcal{R}_{\mathcal{G}}+ \mathcal{R}_{\mathcal{G}}^{'}\leq C_p T^p$ for large enough $T$. This means that $\log \left(\mathcal{R}_{\mathcal{G}}+ \mathcal{R}_{\mathcal{G}}^{'}\right)\leq p\log T+\log C_p$, so 
    \begin{equation*}
    \limsup_{T\rightarrow \infty}\frac{\log \left(\mathcal{R}_{\mathcal{G}}+ \mathcal{R}_{\mathcal{G}}^{'}\right)}{\log T} \leq \limsup_{T\rightarrow \infty}\frac{p\log T+\log C_p}{\log T}=p.
    \end{equation*}
    Since $p$ is an arbitrary positive constant, by taking it to $0$, the transition above holds. Similarly, $\epsilon$ is also an arbitrary positive constant, so by taking it to zero, all the arms for which $j\in \mathcal{O}_{i}^{'}\setminus \mathcal{O}_{i}$ disappear from the sum above, as $\mathrm{KL}(q_{ij},1/2)=0$. This concludes the proof.
\end{proof}

As the final part of this section, we provide a proof for Theorem~\ref{theo:DB Multi-Player Lower Bound}.
\begin{proof}
We use the definition of the regret and Lemma~\ref{lemma:DB Multi-Player Visitations Lower Bound},
\begin{equation*}
    \begin{aligned}
          \mathbb{E}\mathcal{R}_{\mathcal{G},T}(\pi,\nu)&=\mathbb{E}_{\nu}\left[\frac{1}{2}\sum_{i\in [K]}\sum_{j \in [K]\setminus\{i\}}\frac{\Delta_{1i}+\Delta_{1j}}{2}N_{ij}(T)+\sum_{i\in [K]}\Delta_{1i}N_{ii}(T)\right]\\
          &\geq \mathbb{E}_{\nu}\left[\sum_{i,j \in [K]: q_{ij}<1/2}\frac{\Delta_{1i}+\Delta_{1j}}{2}N_{ij}(T)+\sum_{i\in [K]}\Delta_{1i}N_{ii}(T)\right] \\
          & \geq \mathbb{E}_{\nu}\left[\sum_{i \in [K]\setminus \{1\}}\sum_{j \in \mathcal{O}_{i}}\frac{\Delta_{1i}+\Delta_{1j}}{2}N_{ij}(T)\right]\\
          &=\mathbb{E}_{\nu}\left[\sum_{i \in [K]\setminus \{1\}}\sum_{j \in \mathcal{O}_{i}}\frac{\Delta_{1i}+\Delta_{1j}}{2\mathrm{KL}(q_{ij},1/2)}\mathrm{KL}(q_{ij},1/2)N_{ij}(T)\right]\\
          &\geq \sum_{i \in [K]\setminus \{1\}}\min_{j \in \mathcal{O}_{i}}\frac{\Delta_{1i}+\Delta_{1j}}{2\mathrm{KL}(q_{ij},1/2)}\mathbb{E}_{\nu}\left[\sum_{j \in \mathcal{O}_{i}}\mathrm{KL}(q_{ij},1/2)N_{ij}(T)\right].
    \end{aligned}
\end{equation*} 
\end{proof}

\section{FOLLOW YOUR LEADER BLACK BOX ALGORITHM PROOFS}
\label{Appendix B: Follow Your Leader Black Box Algorithm Proofs}
In this section, we prove and elaborate on the various claims from Section\ref{sec: Followers Your Leader Black Box Algorithm}. 
\subsection{A Simple Leader Election Algorithm}
\label{subsec: Leader Election Algorithm}
Since our focus in Theorem~\ref{theorem:FYLBlackBox} is on the asymptotic regime, we narrow our attention to deterministic identified leader election algorithms, denoted as LEAlg, capable of completion within $T_{\mathrm{LE}}$ rounds, irrespective of the overall time horizon $T$. An illustrative example is presented in Algorithm~\ref{alg:Simple Leader Election}, which also finds application in the multiplayer MAB setup discussed in \citet{wang2020optimal}. Our assumption entails that each player is initialized with a unique deterministic ID in the range $[M]$. Subsequently, players share their IDs with neighbors and update their IDs based on the minimum value obtained from their own ID and ones that were received from neighbors. 
\begin{algorithm}[t]
    \DontPrintSemicolon
    \SetAlgoVlined
    \SetKwInOut{Input}{Input}
    \SetKwInOut{Initialize}{Initialize}
    \caption{Simple Leader Election}\label{alg:Simple Leader Election}
    \textbf{Input:} Number of players $M$, communication graph $\mathcal{G}=(\mathcal{V},\mathcal{E})$.\;
    \textbf{Initialize:} $\text{ID}(m)$ - unique ID in $[M]$ for $m \in \mathcal{V}$.\;
    \For{$t=1,\ldots,D+1$}{
        \For{$m=1,\ldots,M$}{
            Send $\text{ID}(m)$ to neighbors.\;
            Receive ID values originating from neighbors $\mathcal{N}(m)$.\;
            $\text{ID}(m)\leftarrow \min\{\text{ID}(m')~|~m'\in\mathcal{N}(m) ~or~ m'=m\}$.\;
        }
    }
\end{algorithm}
Consequently, by the end of the leader election phase, the player originally possessing the minimum ID assumes the role of the leader, while the remaining players become followers as their IDs undergo changes.

For this algorithm, leader election is completed within $T_{\mathrm{LE}}=D+1$ rounds. Additional examples of applicable algorithms can be found in  \citet{casteigts2019deterministic}. 
\subsection{Proof for the Claims in Section~\ref{sec: Followers Your Leader Black Box Algorithm}}
\label{subsec: Proof for the claims in section 2}
We start with a proof for Theorem~\ref{theorem:FYLBlackBox}.
\begin{proof}
    First, we discuss the random regret $\mathcal{R}(T)$. This can be decomposed into four parts:
    \begin{itemize}
        \item \textbf{Leader Election Phase:} For a deterministic $T_{\mathrm{LE}}$ number of rounds, each player draws some arm pair determined by the base algorithm. Taking into account the time required for the leader's CW candidate to reach all the players for the first time after the former's election, we have a contribution of $M(T_{\mathrm{LE}}+D)\Delta_{1\max}$ to the regret bound. 
        \item \textbf{Followers Exploitation Rounds:} For rounds where the followers draw the same arm as the CW candidate held by the leader, regret is incurred only if this arm is not the CW. This results in a regret bound contribution of  $(M-1)\sum_{t=T_{\mathrm{LE}}+1}^{T}\Delta_{1\max}\mathds{1}\{\mathrm{cw}(t) \neq 1\}.$
        \item \textbf{Communication Rounds:} These are rounds in which the arm drawn by followers differs from the CW candidate held by the leader due to communication delay, i.e., rounds where the updated arm information has not yet reached these followers. Given the maximal delay of $D$ rounds, this contributes an additional factor of $MD\Delta_{1\max}$ to the regret bound, as in the worst case a given player draws at round $T$ the leader's candidate from round $T-D$. 
        \item \textbf{Leader's Regret:} The leader draws arms as indicated by the base algorithm, and by only utilizing its own samples, and the resulting contribution to the regret is denoted by $\mathcal{R}_{m_l}(T)$. 
    \end{itemize}
    Using the decomposition above, we bound the random regret as,
    \begin{equation*}
        \mathcal{R}(T) \leq M(T_{\mathrm{LE}}+2D)\Delta_{1\max}+M\Delta_{1\max}\sum_{t=T_{\mathrm{LE}}+1}^{T}\mathds{1}\{\mathrm{cw}(t)\neq 1\}+\mathcal{R}_{m_l}(T)~.
    \end{equation*}
    To complete the proof for the average regret, let us denote $\mathbb{P}$ as the probability measure for the entire multiplayer system, and $\mathbb{P}'$ as the measure for a single-player system containing only the leader operating as indicated by the base algorithm. For any history $\mathcal{H}^{m_l}_{t}$ containing only the leader's draws and rewards, we show that it has the same probability under both measures. To this end, we use an induction principle. This holds trivially for $t=1$, and for $t>1$ we get that since the leader uses the base algorithm without information outside of $\mathcal{H}^{m_l}_{t}$, and since it is elected deterministically by the leader election algorithm, the following holds.
    \begin{equation*}
        \begin{aligned}
            &P(\mathcal{H}^{m_l}_{t+1})=P((i_{m_l}(t+1),j_{m_l}(t+1)),r_{m_l}(t+1),\mathcal{H}^{m_l}_{t})\\
            &=P((i_{m_l}(t+1),j_{m_l}(t+1)),r_{m_l}(t+1)|\mathcal{H}^{m_l}_{t})P(\mathcal{H}^{m_l}_{t})\\
            &=P'((i_{m_l}(t+1),j_{m_l}(t+1)),r_{m_l}(t+1)|\mathcal{H}^{m_l}_{t})P'(\mathcal{H}^{m_l}_{t})=P'(\mathcal{H}^{m_l}_{t+1})~,
        \end{aligned}
    \end{equation*}
    where the third transition follows from the induction assumption and from the fact that the leader follows the base algorithm in both environments. Since neither $\mathcal{R}_{m_l}(T)$ nor $\mathrm{cw}(t)$ contain information outside $\mathcal{H}^{m_l}_{t}$, Assumption \ref{assumption: BlackBoxAlg} leads to,
    \begin{equation*}
        \begin{aligned}
            \mathbb{E}\mathcal{R}(T) &\leq M(T_{\mathrm{LE}}+2D)\Delta_{1\max}+M\Delta_{1\max}\mathbb{E}_{\mathbb{P}}\left[\sum_{t=T_{\mathrm{LE}}+1}^{T}\mathds{1}\{\mathrm{cw}(t)\neq 1\}\right]+\mathbb{E}_{\mathbb{P}}\mathcal{R}_{m_l}(T)\\
            &= M(T_{\mathrm{LE}}+2D)\Delta_{1\max}+M\Delta_{1\max}\mathbb{E}_{\mathbb{P'}}\left[\sum_{t=T_{\mathrm{LE}}+1}^{T}\mathds{1}\{\mathrm{cw}(t)\neq 1\}\right]+\mathbb{E}_{\mathbb{P'}}\mathcal{R}_{m_l}(T)\\
            &\leq M(T_{\mathrm{LE}}+2D)\Delta_{1\max}+M\Delta_{1\max}f(K,T,Q)+g(K,T,Q)~.
        \end{aligned}
    \end{equation*}
    The decomposition above guarantees that communication, once the leader election phase is completed, is only initiated by the leader in rounds where $\mathrm{cw}(t-1) \neq \mathrm{cw}(t)$, and the expected number of such rounds is bounded by $2f$. This completes the proof for the expected regret bound. Additionally, this shows that the expected number of communication rounds is bounded by $T_{\mathrm{LE}}+2f(K,T,Q)$. 
\end{proof}
Next, to prove Corollary~\ref{corollary:FYLRMED} we present the following lemma. 

\begin{lemma}
	\label{lemma: RMED RUCB CW mistake bound}
     For any $\epsilon>0$ and $\alpha>1$, the following holds.
     \begin{enumerate}[label=(\alph*)]
         \item For RUCB \citep{zoghi2014relative} as the base algorithm, define the CW candidate $\mathrm{cw}(t)$ at each round as the hypothesized best arm  $\mathcal{B}(t)$ if it is not empty. Otherwise, use a random arm. Then,
            \begin{equation*}
            \begin{aligned}
                f^{RUCB}(K,T,Q)&=O\left(\frac{K^2 \log K}{\Delta^2_{1\min}}\right)~,\\
                g^{RUCB}(K,T,Q)&=\sum_{i=2}^{K}\frac{4\alpha \log T}{\Delta_{1i}}+O\left(\frac{K^2 \log K}{\Delta^2_{1\min}}\right)~.
            \end{aligned}
        \end{equation*}
         \item For RMED2FH \citep{komiyama2015regret} as the base algorithm, define the CW candidate $\mathrm{cw}(t)$ at each round to be the arm $i^{*}(t)=\argmin I_{i}(t)$, where $I_i(t)$ is the empirical divergence of arm $i$. Then,
         \begin{equation*}
            \begin{aligned}
                f^{RMED2FH}(K,T,Q)&=O\left(K^2\log\log T+K^{2+\epsilon}\right)~,\\
                g^{RMED2FH}(K,T,Q)&=\sum_{i=2}^{K}\min_{j \in \mathcal{O}_{i}}\frac{\Delta_{1i}+\Delta_{1j}}{2\mathrm{KL}(q_{ij},1/2)}\log T\\
                &+O\left(K^2\log\log T+K^{2+\epsilon}+\frac{K\log T}{\log\log T}\right)~.
            \end{aligned}
        \end{equation*}
     \end{enumerate}
 \end{lemma}
\begin{proof}
    We begin with a proof for part 1, utilizing RUCB as a base algorithm. By definition, the bound $g^{RUCB}$ is identical to the expected regret bound of RUCB. Following a proof scheme similar to that presented in \citet{zoghi2014relative}, this results in,
    \begin{equation*}
        g^{RUCB}(K,T,Q) = \left[2\mathcal{D}\log 2\mathcal{D}+2\left(K^{2}\frac{4\alpha-1}{2\alpha-1}\right)^{\frac{1}{2\alpha-1}}\frac{2\alpha-1}{\alpha-1}\right]\Delta_{1\max}+\sum_{i\in [K] / cw}\frac{4\alpha \log T}{\Delta_{1i}}~.
    \end{equation*}
    Here $\mathcal{D}=\sum_{i<j}D_{ij}~$, where $D_{1j}=\frac{4\alpha}{\Delta_{1j}^2}$ for $i=1$ and $D_{ij}=\frac{4\alpha}{\min\{\Delta_{1i}^2,\Delta_{1j}^2\}}$ for $i,j\neq 1$.
    For the bound $f^{RUCB}$, with probability larger than $1-\delta$ it holds that the hypothesized CW $\mathcal{B}(t)$ contains the CW for all $t \geq \hat{T}_{\delta}$. Therefore, 
     \begin{equation*}
         \sum_{t=1}^{T}\mathds{1}\{\mathrm{cw}(t)\neq 1\} \leq \hat{T}_{\delta} \leq 2\mathcal{D}\log 2\mathcal{D} + 2C(\delta)~,
     \end{equation*}
     where we use some notations from the proof scheme in \citet{zoghi2014relative}. By using a similar integration technique to the one utilized in Appendix~\ref{sec:Appendix C: Message Passing RUCB Proofs} to establish a bound for the expected regret given a high probability regret bound, we obtain that,
     \begin{equation*}
         \mathbb{E}\left[\sum_{t=1}^{T}\mathds{1}\{\mathrm{cw}(t)\neq 1\}\right] \leq 2\mathcal{D}\log 2\mathcal{D}+2\left(K^{2}\frac{4\alpha-1}{2\alpha-1}\right)^{\frac{1}{2\alpha-1}}\frac{2\alpha-1}{\alpha-1}~.
     \end{equation*}
     The proof for part one is concluded by observing that  $\mathcal{D}=O(K^2 \log K/\Delta^2_{1\min})$. 

     For the second part, the bound $g^{RMED2FH}$ follows from the expected regret bound provided in \citet{komiyama2015regret}. For the bound $f^{RMED2FH}$, define the event 
     \begin{equation*}
         \mathcal{U}(t):= \cap_{i\in [K]\backslash1}\{\hat{q}_{1,i}(t)>1/2\}~.
     \end{equation*}
     By definition, when $\mathcal{U}(t)$ is true we have $I_{i}(t)>0$ for any arm $i\neq 1$, making $i^{*}(t)=1$ unique with $I_{i^{*}(t)}=0$. Therefore,
     \begin{equation*}
         \sum_{t=1}^{T}\mathds{1}\{\mathrm{cw}(t)\neq 1\} \leq \sum_{t=1}^{T}\mathds{1}\{\mathcal{U}^{c}(t)\}~.
     \end{equation*}
     From Lemma 5 in \citet{komiyama2015regret}, 
     \begin{equation*}
         \mathbb{E}\left[\sum_{t=T_{init}+1}^{T}\mathds{1}\{\mathcal{U}^{c}(t)\}\right] =O\left(K^{2+\epsilon}\right)~,
     \end{equation*}
     for any $\epsilon>0$. Finally, by substituting the value of $T_{init}$ for RMED2FH, it holds that,
     \begin{equation*}
     \begin{aligned}
         f^{RMED2FH}(K,T,Q)=\mathbb{E}\left[\sum_{t=1}^{T}\mathds{1}\{\mathrm{cw}(t)\neq 1\}\right] = O\left(K^{2}\log\log T+K^{2+\epsilon}\right)~.
     \end{aligned}
     \end{equation*}
\end{proof}
Corollary~\ref{corollary:FYLRMED} in the main paper follows directly from the combination of the previous Lemma and Theorem~\ref{theorem:FYLBlackBox}.

\section{MESSAGE-PASSING RUCB PROOFS}
\label{sec:Appendix C: Message Passing RUCB Proofs}
In this section, we elaborate on the proof scheme for the theorem and lemmas presented in Section~\ref{sec: A Fully Distributed Approach} and provide a slightly more general formulation that allows for more flexibility in the choice of constants. 

For clarity, it is important to note that the UCB indices calculated by each player $m$ for all $i \neq j$ are expressed as follows.
\begin{equation*}
    u_{ij}^{m}(t) := \frac{\tilde{w}^{m}_{ij}(t-1)}{\tilde{N}^{m}_{ij}(t-1)}+\sqrt{\frac{\alpha\log t}{\tilde{N}^{m}_{ij}(t-1)}}~.
\end{equation*}
Using this notation, at the beginning of round $t$, players employ $\tilde{w}^{m}_{ij}(t-1),\tilde{N}^{m}_{ij}(t-1)$ to update the UCB terms. The values of $\tilde{w}^{m}_{ij}(t),\tilde{N}^{m}_{ij}(t)$ are then updated at the end of the round after the communication phase is concluded. 

In the proof scheme presented in this section, we utilize a different probability space compared to the one defined in Section\ref{Appendix A: Lower Bound Proof}. We introduce a collection of independent Bernoulli random variables $(q_{ij}^{m}(t))$ for $t\in [T],i,j \in [K], m \in [M]$, satisfying $\mathbb{E}q_{ij}^{m}(t)=q_{ij}~$. Caratheodory's extension theorem allows us to define this collection of random variables on the probability space $(\Omega,\mathcal{F})=\left(\mathbb{R}^{TMK^{2}},\mathcal{B}\left(\mathbb{R}^{TMK^{2}}\right)\right)$. Under this model, the reward $r_{m}(t)$ collected by player $m$ at round $t$ is represented as $r_{m}(t)=q^{m}_{i_m(t)j_m(t)}(t)~$. By defining the filtration $\mathcal{F}_t=\sigma\left(q_{ij}^{m}(\tau)\right)_{\tau\leq t}$, we ensure that $i_m(t),j_m(t) \in \mathcal{F}_{t-1}$ and $r_{m}(t) \in \mathcal{F}_{t}~$. Subsequently, we define $r_{m}^{ij}(t)$ as the reward obtained by player $m$, given that the pair $(i,j)$ was drawn at round $t$. Otherwise, it is defined to be $0$.
\begin{equation*}
    r_{ij}^{m}(t) = q_{ij}^{m}(t)\mathds{1}\{(i_m(t),j_m(t)) =(i,j)\}+(1-q_{ji}^{m}(t))\mathds{1}\{(i_m(t),j_m(t)) =(j,i)\}~,
\end{equation*}
In addition, define the LCB terms for all $i,j,m$ as,
\begin{equation*}
    l_{ij}^{m}(t) = 1 - u_{ij}^{m}(t)~.
\end{equation*}
We begin by presenting a concentration bound in the following lemma. 
\begin{lemma}
	\label{lemma:MP concentration bound}
	The following concentration bounds hold for all $i,j,m,t>0,\eta>1$ and $\delta>0$.
 \begin{equation*}
     \begin{aligned}
         P\left(q_{ij}< l_{ij}^{m}(t)\right) &< \frac{\log \left(d_m(\mathcal{G}_{\gamma})+1\right)t}{t^{2\alpha \eta^{-1/2}}\log \eta}\\
         P\left(q_{ij}> u_{ij}^{m}(t)\right) &< \frac{\log \left(d_m(\mathcal{G}_{\gamma})+1\right)t}{t^{2\alpha \eta^{-1/2}}\log \eta}~, 
     \end{aligned}
 \end{equation*}
 where $\mathcal{G}_{\gamma}$ stands for the $\gamma$-power of the graph $\mathcal{G}$, and $d_m(\cdot)$ is the degree of node $m$ on the graph.
 \end{lemma}
 \begin{proof}
    For $i=j$, this holds by definition. Going forward, we concentrate on the case $i \neq j$. For simplicity, let us denote
    \begin{equation*}
        \mathds{1}^{m,m'}(t,\tau):=\mathds{1}\{d(m,m')\leq \min(\gamma,t-\tau+1)\}~,
    \end{equation*}
    and
    \begin{equation*}
        \chi_{ij}^{m,m'}(t,\tau):=\mathds{1}\{(i_{m'}(\tau),j_{m'}(\tau))=(i,j)\}\mathds{1}^{m,m'}(t,\tau)~.
    \end{equation*}
    This indicator determines whether the arm pair $(i,j)$ was drawn by player $m'$ at round $\tau$ and whether it is included in the statistics of player $m$ at time $t$. The latter condition is satisfied only if the distance between the nodes on the graph is smaller than $\gamma$ and if the time difference between $t$ and the drawing time $\tau$ is greater than the time it takes to traverse the graph ($d(m,m')-1$). Next, for $\tau \leq t$ define,
     \begin{equation*}
         \tilde{r}_{ij}^{m,m'}(t,\tau):=q_{ij}^{m'}(\tau)\chi_{ij}^{m,m'}(t,\tau)+(1-q_{ji}^{m'}(\tau))\chi_{ji}^{m,m'}(t,\tau)~,
     \end{equation*}
     and,
     \begin{equation*}
         Y_{ij}^{m}(t,\tau):= \sum_{m'=1}^{M}\left(\tilde{r}_{ij}^{m,m'}(t,\tau)-\mathbb{E}\left[\tilde{r}_{ij}^{m,m'}(t,\tau)|\mathcal{F}_{\tau-1}\right]\right)~.
     \end{equation*}
    For any player $m$ and arm pair $(i,j)$, $Y_{ij}^{m}(t,\tau)$ represents the contribution at round $t$ of conditional centralized rewards drawn at round $\tau$ from all players from which the message can be received at the current round. Since the first indicator in the definition of $\chi_{ij}^{m,m'}(t,\tau)$ only depends on draws at round $\tau$ and the second indicator is deterministic, it holds that $\chi_{ij}^{m,m'}(t,\tau) \in \mathcal{F}_{\tau-1}$. Therefore, we can decompose $Y_{ij}^{m}(t,\tau)$ as follows.
    \begin{equation*}
         Y_{ij}^{m}(t,\tau)= \sum_{m'=1}^{M}\left((q_{ij}^{m'}(\tau)-q_{ij})\chi_{ij}^{m,m'}(t,\tau)+(1-q_{ji}^{m'}(\tau)-q_{ij})\chi_{ji}^{m,m'}(t,\tau)\right)~.
     \end{equation*}
    By taking an expected value, it holds that for any $\lambda \in \mathbb{R}$,
    \begin{equation*}
        \begin{aligned}
           & \mathbb{E}\left[\exp\left(\lambda Y_{ij}^{m}(t,\tau)\right)|\mathcal{F}_{\tau-1}\right] \\
            &= \mathbb{E}\left[\exp\left(\lambda \sum_{m'=1}^{M}\left((q_{ij}^{m'}(\tau)-q_{ij})\chi_{ij}^{m,m'}(t,\tau)+(1-q_{ji}^{m'}(\tau)-q_{ij})\chi_{ji}^{m,m'}(t,\tau)\right)\right)\Bigg|\mathcal{F}_{\tau-1}\right]\\
            &= \prod_{m'=1}^{M}\mathbb{E}\left[\exp\left(\lambda \left((q_{ij}^{m'}(\tau)-q_{ij})\chi_{ij}^{m,m'}(t,\tau)+(1-q_{ji}^{m'}(\tau)-q_{ij})\chi_{ji}^{m,m'}(t,\tau)\right)\right)\Big|\mathcal{F}_{\tau-1}\right]\\
            & = \prod_{m'=1}^{M}\mathbb{E}\left[\exp\left(\lambda (q_{ij}^{m'}(\tau)-q_{ij})\chi_{ij}^{m,m'}(t,\tau)\right)\big|\mathcal{F}_{\tau-1}\right]\\
            &\times\prod_{m'=1}^{M}\mathbb{E}\left[\exp\left(\lambda (1-q_{ji}^{m'}(\tau)-q_{ij})\chi_{ji}^{m,m'}(t,\tau)\right)\big|\mathcal{F}_{\tau-1}\right]\\
            &\leq \prod_{m'=1}^{M}\exp\left(\frac{\lambda^2}{8}\chi_{ij}^{m,m'}(t,\tau)\right)\exp\left(\frac{\lambda^2}{8}\chi_{ji}^{m,m'}(t,\tau)\right)\\
            &= \prod_{m'=1}^{M}\exp\left(\frac{\lambda^2}{8}\left(\chi_{ij}^{m,m'}(t,\tau)+\chi_{ji}^{m,m'}(t,\tau)\right)\right)\\
            & = \exp\left(\frac{\lambda^2}{8}\sum_{m'=1}^{M}\zeta_{ij}^{m,m'}(t,\tau)\right)~,
        \end{aligned}
    \end{equation*}
    where 
    \begin{equation*}
        \zeta_{ij}^{m,m'}(t,\tau) := \zeta^{m'}_{ij}(\tau)\mathds{1}^{m,m'}(t,\tau)~,
    \end{equation*}
    and $\zeta^{m}_{ij}(t)$ is an indicator specifying whether the pair $(i,j)$ was drawn by player $m$ at round $t$ in any order,
    \begin{equation*}
        \zeta_{ij}^{m}(t):= \mathds{1}\{(i_m(t),j_m(t)) \in \{(i,j),(j,i)\}\}~.
    \end{equation*}
    In the second transition above, we leverage the independence of $q_{ij}^{m}(t)$ across different players and arms, and the fact that all other terms are determined by conditioning on $\mathcal{F}_{\tau-1}$. The third transition follows a similar logic. For the fourth transition, we use the fact that both $q_{ij}^{m}(t)-q_{ij}$ and $1-q_{ji}^{m}(t)-q_{ij}$ are centered Bernoulli random variables, and as such, they are also $1/2-$sub-Gaussian. The inequality above implies that,
    \begin{equation*}
        \mathbb{E}\left[\exp\left(\lambda Y_{ij}^{m}(t,\tau)-\frac{\lambda^2}{8}\sum_{m'=1}^{M}\zeta_{ij}^{m,m'}(t,\tau)\right)|\mathcal{F}_{\tau-1}\right] \leq 1~.
    \end{equation*}
     Note that the visitation counter utilized by players can be decomposed as the following sum of contributions from all players and prior rounds.
     \begin{equation*}
         \begin{aligned}
             \tilde{N}_{ij}^{m}(t) = \sum_{\tau=1}^{t}\sum_{m'=1}^{M}\zeta_{ij}^{m,m'}(t,\tau)~.
         \end{aligned}
     \end{equation*}
     Next, define 
     \begin{equation*}
         Z_{ij}^{m}(t) := \sum_{\tau=1}^{t}Y_{ij}^{m}(t,\tau)~.
     \end{equation*}
     It holds that, 
     \begin{equation*}
         \begin{aligned}
             Y_{ij}^{m}(t,\tau) &= \sum_{m'=1}^{M}\left(r_{ij}^{m'}(\tau)\mathds{1}^{m,m'}(t,\tau)-q_{ij}\zeta^{m'}_{ij}(\tau)\mathds{1}^{m,m'}(t,\tau)\right)\\
             &= \sum_{m'=1}^{M}r_{ij}^{m'}(\tau)\mathds{1}^{m,m'}(t,\tau)- q_{ij}\sum_{m'=1}^{M}\zeta^{m,m'}_{ij}(t,\tau)~,
         \end{aligned}
     \end{equation*}
     meaning that $Z_{ij}^{m}(t)$ is a random variable representing the centered number of wins,
     \begin{equation*}
         Z_{ij}^{m}(t) = \tilde{w}^{m}_{ij}(t)-q_{ij}\tilde{N}^{m}_{ij}(t)~.
     \end{equation*}
     Employing the previous inequality, it holds that,
\begin{equation*}
    \begin{aligned}
        & \mathbb{E}\left[\exp\left(\lambda Z_{ij}^{m}(t) - \frac{\lambda^2}{8} \tilde{N}_{ij}^{m}(t)\right) \Bigg| \mathcal{F}_{t-1} \right] \\
        &= \mathbb{E}\left[\exp\left(\lambda Y_{ij}^{m}(t,t) - \frac{\lambda^2}{8} \sum_{m'=1}^{M} \zeta_{ij}^{m,m'}(t,t) \right. \right. \\
        &\quad \left. \left. + \lambda \sum_{\tau=1}^{t-1} Y_{ij}^{m}(t,\tau) - \frac{\lambda^2}{8} \sum_{\tau=1}^{t-1} \sum_{m'=1}^{M} \zeta_{ij}^{m,m'}(t,\tau) \right) \Bigg| \mathcal{F}_{t-1} \right] \\
        &= \mathbb{E}\left[\exp\left(\lambda Y_{ij}^{m}(t,t) - \frac{\lambda^2}{8} \sum_{m'=1}^{M} \zeta_{ij}^{m,m'}(t,t) \right) \Bigg| \mathcal{F}_{t-1} \right] \\
        &\quad \times \exp\left(\lambda \sum_{\tau=1}^{t-1} Y_{ij}^{m}(t,\tau) - \frac{\lambda^2}{8} \sum_{\tau=1}^{t-1} \sum_{m'=1}^{M} \zeta_{ij}^{m,m'}(t,\tau) \right) \\
        &\leq \exp\left(\lambda \sum_{\tau=1}^{t-1} Y_{ij}^{m}(t,\tau) - \frac{\lambda^2}{8} \sum_{\tau=1}^{t-1} \sum_{m'=1}^{M} \zeta_{ij}^{m,m'}(t,\tau) \right)~,
    \end{aligned}
\end{equation*}
     where the second transition follows the fact that samples collected before time $t$ are determined by $\mathcal{F}_{t-1}$. By extending this inequality further in time and applying the tower rule, we obtain the following.
     \begin{equation*}
         \mathbb{E}\left[\exp\left(\lambda Z_{ij}^{m}(t)-\frac{\lambda^2}{8}\tilde{N}_{ij}^{m}(t)\right)\right] \leq 1~. 
     \end{equation*}
     Next, from the Markov inequality it holds that for any $\theta,\kappa,\lambda \in \mathbb{R}$,
     \begin{equation*}
         e^{-2\kappa\theta } \geq P\left(\exp\left(\lambda Z_{ij}^{m}(t)-\frac{\lambda^2}{8}\tilde{N}_{ij}^{m}(t)\right) \geq e^{2\kappa\theta}\right)=P\left(\lambda Z_{ij}^{m}(t)-\frac{\lambda^2}{8}\tilde{N}_{ij}^{m}(t) \geq 2\kappa\theta\right)~.
     \end{equation*}
     While this is valid for any $\lambda\in \mathbb{R}$, dividing the subsequent steps into segments for $\lambda>0$ and $\lambda<0$ will facilitate the proofs of the first and second concentration bounds, respectively. Given the similarity between the two proofs, we concentrate on the case $\lambda>0$ in the following. It is evident that,
     \begin{equation*}
         P\left(\frac{Z_{ij}^{m}(t)}{\sqrt{\tilde{N}_{ij}^{m}(t)}} \geq \frac{2\kappa\theta}{\lambda \sqrt{\tilde{N}_{ij}^{m}(t)}}+\frac{\lambda}{8}\sqrt{\tilde{N}_{ij}^{m}(t)}\right) \leq e^{-2\kappa\theta }~.
     \end{equation*}
     Next, we note that for any pair of arms $(i,j)$, player $m$ can receive no more than $d_{m}(\mathcal{G}_{\gamma})$ samples at each round from other players. Hence, $1 \leq \tilde{N}_{ij}^{m}(t) \leq \left(1+d_{m}(\mathcal{G}_{\gamma})\right)t$. Utilizing this fact, for any $\eta>1$, it holds that $1 \leq \tilde{N}_{ij}^{m}(t) \leq \eta^{D_t}$ for 
     \begin{equation*}
         D_{t}:= \frac{\log \left(1+d_{m}(\mathcal{G}_{\gamma})\right)t}{\log \eta}~.
     \end{equation*}
     This enables us to address the randomness in $Z_{ij}^{m}(t),\tilde{N}_{ij}^{m}(t)$ by employing a peeling argument and partitioning the potential values of $\tilde{N}_{ij}^{m}(t)$ into intervals $[\eta^{l-1},\eta^{l}]$ for $1 \leq l \leq D_t$. For this purpose, we introduce
     \begin{equation*}
         \lambda_l:= 4\sqrt{\frac{\kappa \theta}{\eta^{l-0.5}}}~,\kappa = \eta^{-\frac{1}{2}}~.
     \end{equation*}
     For $\tilde{N}_{ij}^{m}(t) \in [\eta^{l-1},\eta^{l}]$ it holds that,
     \begin{equation*}
         \frac{2\kappa\theta}{\lambda_l \sqrt{\tilde{N}_{ij}^{m}(t)}}+\frac{\lambda_l}{8}\sqrt{\tilde{N}_{ij}^{m}(t)} = \frac{\sqrt{\kappa \theta}}{2}\left(\sqrt{\frac{\eta^{l-0.5}}{\tilde{N}_{ij}^{m}(t)}}+\sqrt{\frac{\tilde{N}_{ij}^{m}(t)}{\eta^{l-0.5}}}\right) \leq \sqrt{\theta}~,
     \end{equation*}
     and by summing over the different intervals,
     \begin{equation*}
         \begin{aligned}
             P\left(\frac{Z_{ij}^{m}(t)}{\sqrt{\tilde{N}_{ij}^{m}(t)}} \geq \sqrt{\theta}\right)&\leq \sum_{l=1}^{D_t}P\left(\frac{Z_{ij}^{m}(t)}{\sqrt{\tilde{N}_{ij}^{m}(t)}} \geq \sqrt{\theta},\tilde{N}_{ij}^{m}(t) \in [\eta^{l-1},\eta^{l}]\right) \\
             &\leq \sum_{l=1}^{D_t}P\left(\frac{Z_{ij}^{m}(t)}{\sqrt{\tilde{N}_{ij}^{m}(t)}} \geq \frac{2\kappa\theta}{\lambda_l \sqrt{\tilde{N}_{ij}^{m}(t)}}+\frac{\lambda_l}{8}\sqrt{\tilde{N}_{ij}^{m}(t)},~\tilde{N}_{ij}^{m}(t) \in [\eta^{l-1},\eta^{l}]\right)\\
             &\leq \sum_{l=1}^{D_t}P\left(\frac{Z_{ij}^{m}(t)}{\sqrt{\tilde{N}_{ij}^{m}(t)}} \geq \frac{2\kappa\theta}{\lambda_l \sqrt{\tilde{N}_{ij}^{m}(t)}}+\frac{\lambda_l}{8}\sqrt{\tilde{N}_{ij}^{m}(t)}\right)\\
             &\leq \sum_{l=1}^{D_t}e^{-2\kappa \theta} = \frac{\log \left(1+d_{m}(\mathcal{G}_{\gamma})\right)t}{\log \eta}e^{-2\kappa \theta}~.
         \end{aligned}
     \end{equation*}
     To conclude, substitute $\theta = \alpha \log t$,
     \begin{equation*}
         \begin{aligned}
             P\left(q_{ij}\leq l_{ij}^{m}(t+1)\right) &= P\left(\frac{\tilde{w}^{m}_{ij}(t)}{\tilde{N}^{m}_{ij}(t)}-q_{ij}\geq \sqrt{\frac{\alpha \log t}{\tilde{N}^{m}_{ij}(t)}}\right)\\
             &= P\left(\frac{Z^{m}_{ij}(t)}{\tilde{N}^{m}_{ij}(t)}\geq \sqrt{\frac{\alpha \log t}{\tilde{N}^{m}_{ij}(t)}}\right) \leq \frac{\left(1+d_{m}(\mathcal{G}_{\gamma})\right)(t+1)}{\log \eta}\frac{1}{t^{2\alpha \eta^{-1/2}}}~.
         \end{aligned}
     \end{equation*}
     The concentration bound in the theorem results from replacing $t+1$ with $t$. 
 \end{proof}
 The next lemma demonstrates that with high probability, $q_{ij} \leq u_{ij}^{m}(t)$ for all players, arm pairs, and rounds that are sufficiently large.
\begin{lemma}
	\label{lemma:MP concentration bound union complicated}
	For any $\epsilon>0, \delta>0$, and $\alpha$ such that $\alpha> 0.5\sqrt{\eta}(1+\epsilon)$, define
 \begin{equation*}
     T_\epsilon = \begin{cases*}
      \left(\frac{-W_{-1}(-\epsilon)}{\epsilon}\right)^{\frac{1}{\epsilon}} & if $\epsilon \in (0,e^{-1})$ \\
      0 & if $\epsilon \geq e^{-1}$~,
    \end{cases*}
 \end{equation*}
 where $W_{-1}(\cdot)$ is the lower part of the Lambert function. Additionally, define
 \begin{equation*}
 \label{eq:MP_C_delta}
     C_{\epsilon}(\delta):=\max\left(e^{\frac{1}{\epsilon}},T_\epsilon,\left(\frac{MK^2\left(3+2\log \left(d_{\max}^{\gamma}+1\right)\right)}{\min\{1,2\alpha \eta^{-1/2}-1\}\delta \log \eta}\right)^{\frac{1}{2\alpha \eta^{-1/2}-1-\epsilon}}\right)~,
 \end{equation*}
 where $d_{\max}^{\gamma}:=\max_m\left(d_m(\mathcal{G}_{\gamma})\right)~$. UCB terms $u_{ij}^{m}(t)$ satisfy,
 \begin{equation*}
     P\left(\forall t\geq C_{\epsilon}(\delta), i, j,m:~q_{ij} \leq u_{ij}^{m}(t)\right) > 1-\delta~.
 \end{equation*}
 \end{lemma}
 \begin{proof}
     Utilizing Lemma~\ref{lemma:MP concentration bound}, it holds that,
     \begin{equation*}
         \begin{aligned}
             &P\left(\exists t\geq C_{\epsilon}(\delta), i,j,m:~q_{ij} > u_{ij}^{m}(t)\right) \\
             &\leq MK^{2}\sum_{t=C_{\epsilon}(\delta)}^{\infty}\frac{\log \left(d_{\max}^{\gamma}+1\right)t}{t^{2\alpha \eta^{-1/2}}\log \eta}\\
             &\leq \frac{MK^2}{\log \eta}\sum_{t=C_{\epsilon}(\delta)}^{\infty}\left(\frac{\log t}{t^{2\alpha \eta^{-1/2}}}+\frac{\log \left(d_{\max}^{\gamma}+1\right)}{t^{2\alpha \eta^{-1/2}}}\right)\\
             &\leq \frac{MK^2}{\log \eta}\left[\frac{\log C_{\epsilon}(\delta)}{C_{\epsilon}(\delta)^{2\alpha \eta^{-1/2}}}+\frac{\log \left(d_{\max}^{\gamma}+1\right)}{C_{\epsilon}(\delta)^{2\alpha \eta^{-1/2}}}+\int_{C_{\epsilon}(\delta)}^{\infty}\left(\frac{\log t}{t^{2\alpha \eta^{-1/2}}}+\frac{\log \left(d_{\max}^{\gamma}+1\right)}{t^{2\alpha \eta^{-1/2}}}\right)dt\right]~,
         \end{aligned}
     \end{equation*}
     where we took advantage of the fact that for $t>C_{\epsilon}(\delta)>e$, the functions $\log t/t^{2\alpha \eta^{-1/2}}$  and $1/t^{2\alpha \eta^{-1/2}}$ are decreasing. Calculating the integrals above,
    \begin{equation*}
        \begin{aligned}
            &P\left(\exists t\geq C_{\epsilon}(\delta), i,j,m:~q_{ij} > u_{ij}^{m}(t)\right) \\
            &\leq \frac{MK^2}{\log \eta}\left[\frac{\log C_{\epsilon}(\delta)}{C_{\epsilon}(\delta)^{2\alpha \eta^{-1/2}}}+\frac{\log \left(d_{\max}^{\gamma}+1\right)}{C_{\epsilon}(\delta)^{2\alpha \eta^{-1/2}}}+\right.  \\
            &\quad\quad\quad  +\left. \frac{(2\alpha \eta^{-1/2}-1)\log C_{\epsilon}(\delta) +1}{(2\alpha \eta^{-1/2}-1)^{2}C_{\epsilon}(\delta)^{2\alpha \eta^{-1/2}-1}}+\frac{\log \left(d_{\max}^{\gamma}+1\right)}{(2\alpha \eta^{-1/2}-1)C_{\epsilon}(\delta)^{2\alpha \eta^{-1/2}-1}}\right]~.
        \end{aligned}
    \end{equation*}
    For clarity, we denote $\overline{T}:=C_{\epsilon}(\delta)$ in the remainder of this proof. To evaluate the right-hand side above, we are interested in the approximation $\log \overline{T} \leq \overline{T}^{\epsilon}~$ for some $\epsilon>0$. For $\epsilon \geq e^{-1}$, this holds for all $\overline{T}>0$, and for $\epsilon \in (0,e^{-1})$ this is true for $\overline{T} \geq \left(-W_{-1}(-\epsilon)/\epsilon\right)^{1/\epsilon}~$, where $W_{-1}(\cdot)$ stands for the lower part of the Lambert function. Therefore, when this inequality holds,
    \begin{equation*}
        \begin{aligned}
            &P\left(\exists t\geq C_{\epsilon}(\delta), i,j,m:~q_{ij} > u_{ij}^{m}(t)\right) \\
            &\leq \frac{MK^2}{\log \eta}\left[\frac{\log \overline{T}}{\overline{T}^{2\alpha \eta^{-1/2}}}+\frac{\log \left(d_{\max}^{\gamma}+1\right)}{\overline{T}^{2\alpha \eta^{-1/2}}}\right. \\
            &\quad\quad\quad \left. +\frac{2\log \overline{T}}{(2\alpha \eta^{-1/2}-1)\overline{T}^{2\alpha \eta^{-1/2}-1}}+\frac{\log \left(d_{\max}^{\gamma}+1\right)}{(2\alpha \eta^{-1/2}-1)\overline{T}^{2\alpha \eta^{-1/2}-1}}\right]\\
            &\leq \frac{MK^2}{\log \eta}\left[\frac{1}{\overline{T}^{2\alpha \eta^{-1/2}-1-\epsilon}}+\frac{\log \left(d_{\max}^{\gamma}+1\right)}{\overline{T}^{2\alpha \eta^{-1/2}-1-\epsilon}}\right. \\
            &\quad\quad\quad \left. +\frac{2}{(2\alpha \eta^{-1/2}-1)\overline{T}^{2\alpha \eta^{-1/2}-1-\epsilon}}+\frac{\log \left(d_{\max}^{\gamma}+1\right)}{(2\alpha \eta^{-1/2}-1)\overline{T}^{2\alpha \eta^{-1/2}-1-\epsilon}}\right]\\
            &\leq \frac{MK^2\left(3+2\log \left(d_{\max}^{\gamma}+1\right)\right)}{\min\{1,2\alpha \eta^{-1/2}-1\}\log \eta}\frac{1}{\overline{T}^{2\alpha \eta^{-1/2}-1-\epsilon}}~,
        \end{aligned}
    \end{equation*}
    where we also used the fact that $\overline{T}>e^{1/\epsilon}$. By requiring the right-hand side to be smaller than $\delta$, we have,
    \begin{equation*}
        C_{\epsilon}(\delta) \geq \left(\frac{MK^2\left(3+2\log \left(d_{\max}^{\gamma}+1\right)\right)}{\min\{1,2\alpha \eta^{-1/2}-1\}\delta \log \eta}\right)^{\frac{1}{2\alpha \eta^{-1/2}-1-\epsilon}}~.
    \end{equation*}
    The proof concludes by combining the requirements on $C_{\epsilon}(\delta)$ mentioned above.
 \end{proof}
Next, we show that with high probability, players in a clique cannot visit arm pairs different from $(1,1)$ for too many rounds. For brevity, we define,
\begin{equation*}
     D_{ij}:= \begin{cases*}
      4\alpha/\overline{\Delta}_{ij}^{2} & if $i \neq j$ \\
      0 & if $i=j$~.
    \end{cases*}     
 \end{equation*}
\begin{lemma}
	\label{lemma:MP visitation bound complicated}
	For any $\epsilon>0, \delta>0$,$\gamma'\leq \gamma$, $T_0 \geq C_{\epsilon}(\delta)$ and $\alpha$ such that $\alpha> 0.5\sqrt{\eta}(1+\epsilon)$, let $\mathcal{C}(\mathcal{G}_{\gamma'})$ be the set of all cliques in $\mathcal{G}_{\gamma'}$. The number of visitations satisfies,
 \begin{equation*}
     P\left(\exists t>T_0,(i,j)\neq (1,1),\gamma'\leq\gamma, C \in \mathcal{C}(\mathcal{G}_{\gamma'}):~\sum_{m \in C}N_{ij}^{m,T_0}(t)>D_{ij}\log t+(\gamma' +2)|C|\right)< \delta~.
 \end{equation*}
 \end{lemma}
\begin{proof}
     For the remainder of this proof, we assume that the good event $\{\forall t\geq C_{\epsilon}(\delta), i,j,m:~q_{ij} \leq u_{ij}^{m}(t)\}$ holds, which occurs with probability $1-\delta$ according to Lemma~\ref{lemma:MP concentration bound union complicated}. First, for draws of identical arms $(i,i)\neq (1,1)$ during rounds $t>C_{\epsilon}(\delta)$, the RUCB decision rule stipulates that player $m$ can only draw arm $i$ as both the first and the second arm if $u^{m}_{ji}(t)<0.5$ for all $j \neq i$, since it holds that $u^{m}_{ii}(t)=0.5$. However, given the good event, $u^{m}_{1i}(t)\geq q_{1i}>0.5$, which is a contradiction. Hence, pairs of identical arms, apart from $(1,1)$, cannot be drawn under this event. Consequently, for any $C \in \mathcal{C}(\mathcal{G}_{\gamma'})$ it holds that $\sum_{m\in C}N_{ii}^{m,T_0}(t)=0<D_{ii}\log t+(\gamma' +1)|C|$.

     For pairs of non-identical arms $i \neq j$ with some given $C \in \mathcal{C}(\mathcal{G}_{\gamma'})$, let us assume that these arms have been compared for a sufficiently large number of rounds between $T_0$ and $t$, such that $\sum_{m\in C}N_{ij}^{m,T_0}(t-1)>D_{ij}\log t+(\gamma' +1)|C|$. Note that this also dictates $t \geq \gamma'+1+T_0$. Let $s$ denote the last round in which this pair was drawn by any player in $C$, such that $s\geq T_0+\gamma'+1$ \footnote{During round $s$, several players may have drawn this pair.}. Next, Consider several cases for the draw at round $s$, and notice that Algorithm~\ref{alg:MPRUCB} dictates that the first arm for each player $m$ is always chosen from the winners set $\mathcal{C}^{m}$, which necessarily contains the CW and is thus non-empty.
     \begin{itemize}
         \item For $i,j \neq 1$, assume that $i_m(s)=i,j_m(s)=j$. It holds that $u^{m}_{ij}(s)\geq 0.5$, and $l_{ij}^{m}(s):=1-u_{ji}^{m}(s)\leq 1-u_{1i}^{m}(s)\leq 1-q_{1i}(s)$. Therefore, $u_{ij}^{m}(s)-l_{ij}^{m}(s)\geq \Delta_{1i}$.
         \item For $i,j \neq 1$ and $i_m(s)=j,j_m(s)=i$, a similar derivation reveals that $u_{ji}^{m}(s)-l_{ji}^{m}(s)\geq \Delta_{1j}$.
         \item For $i_m(s)=1,j_m(s)=i$ to be drawn, it holds that $u_{1i}^{m}(s)\geq q_{1i}$ and $l_{1i}^{m}(s)=1-u_{i1}^{m}(s)\leq 0.5$. Hence, $u_{1i}^{m}(s)-l_{1i}^{m}(s)\geq \Delta_{1i}$.
         \item For $i_m(s)=i,j_m(s)=1$ to be drawn, it must happen that $u_{i1}^{m}(s)\geq 0.5$ and $l_{i1}^{m}(s)=1-u_{1i}^{m}(s)\leq 1-q_{1i}$. Therefore, it again holds that $u_{i1}^{m}(s)-l_{i1}^{m}(s)\geq \Delta_{1i}$.
     \end{itemize}
     To conclude, for $i \neq j$, a draw of the pair $(i,j)$ at round $s$ leads to $u_{ij}^{m}(s)-l_{ij}^{m}(s) \geq \min\{\Delta_{1i},\Delta_{1j}\}$. Similarly, when one of the arms in the pair is the $CW$, it holds that $u_{i1}^{m}(s)-l_{i1}^{m}(s)\geq \Delta_{1i}$. On the other hand, by definition,
     \begin{equation*}
         \begin{aligned}
             u_{ij}^{m}(s) - l_{ij}^{m}(s) = u_{ij}^{m}(s) + u_{ji}^{m}(s)-1=2\sqrt{\frac{\alpha \log s}{\tilde{N}^{m}_{ij}(s-1)}}~.
         \end{aligned}
     \end{equation*}
     Utilizing the assumption about the number of visits, it holds that,
     \begin{equation*}
         \begin{aligned}
             &\tilde{N}^{m}_{ij}(s-1) \geq \tilde{N}^{m,T_0}_{ij}(s-1)\\
             &\geq \sum_{m \in C}N_{ij}^{m,T_0}(s-\gamma'-1)\geq \sum_{m \in C}N_{ij}^{m,T_0}(s-1)-\gamma'|C|\\
             &> D_{ij}\log t +(\gamma'+1)|C| -|C|-\gamma'|C|\\
             &=D_{ij}\log t~,
         \end{aligned}
     \end{equation*}
    where in the second transition, we leverage the fact that after $\gamma'$ rounds, player $m$ will have received all information available in $C$ in the current round. The third transition takes advantage of the restriction that each player in the clique can only draw at most one pair at a round. Finally, the fourth transition relies on our assumption regarding the number of visits, along with the knowledge that $s$ was the last round the arm pair was drawn by any player until round $t$. Plugging this above,
    \begin{equation*}
        u_{ij}^{m}(s) - l_{ij}^{m}(s) < 2\sqrt{\frac{\alpha}{D_{ij}}}=\overline{\Delta}_{ij}~.
    \end{equation*}
     By the definition of $\overline{\Delta}_{ij}$, this is a contradiction. Therefore, under the good event $\sum_{m\in C}N_{ij}^{m,T_0}(t-1)\leq D_{ij}\log t+(\gamma' +1)|C|$ for all $t>T_0$, $(i,j) \neq (1,1), C \in \mathcal{C}(\mathcal{G}_{\gamma'})$. Hence, $\sum_{m\in C}N_{ij}^{m,T_0}(t)\leq D_{ij}\log t+(\gamma' +2)|C|$, with probability $1-\delta$.
 \end{proof}
     
Next, we demonstrate how the proof to Lemma~\ref{lemma:MP visitation bound} in the main paper follows from the previous lemma.
\begin{proof}
    Take $\eta=1.3$ and $\epsilon = e^{-1}$. For these values, it holds that $T_{\epsilon}=0$, and by demanding $\alpha>1.2$ we have $2\alpha \eta^{-1/2}-1>1$, which makes the requirement $\geq e^{1/\epsilon}$ in the proof to Lemma~\ref{lemma:MP concentration bound union complicated} irrelevant. By combining all of the above, we have that,
     \begin{equation*}
     C(\delta):=\left(\frac{4MK^2\left(3+2\log \left(d_{\max}^{\gamma}+1\right)\right)}{\delta }\right)^{\frac{1}{1.7\alpha -1.4}}~
 \end{equation*}
 is sufficient for the concentration bound in Lemma~\ref{lemma:MP concentration bound union complicated} to hold. By utilizing these in Lemma~\ref{lemma:MP visitation bound complicated}, we derive the version found in the main paper. 
\end{proof}
\begin{lemma}
	\label{lemma:MP T_delta definition}     
     For any $\epsilon>0, \delta>0$,$a>0$, $\gamma'\leq \gamma$ and $\alpha$ such that $\alpha> 0.5\sqrt{\eta}(1+\epsilon)$, let $\hat{T}_{\epsilon}(\delta)$ be the smallest round satisfying
     \begin{equation*}
         \hat{T}_{\epsilon}(\delta) > h(K,\gamma',T_0)+a\log \hat{T}_{\epsilon}(\delta)~,
     \end{equation*}
     where $h(\cdot)$ is independent of $\hat{T}_{\epsilon}(\delta)$. Then, $\hat{T}_{\epsilon}(\delta)$ exists, and it holds that,
     \begin{equation*}
         \hat{T}_{\epsilon}(\delta) \leq 2h(K,\gamma',T_0)+2a\log (2a)~.
     \end{equation*}
 \end{lemma}
 
 \begin{proof}
      To prove the lemma, it is sufficient to find a value of $\tau$ such that $\tau > h(K,\gamma',T_0)+a\log \tau~$. By the definition of $\hat{T}_{\epsilon}(\delta)$, this means that $\hat{T}_{\epsilon}(\delta)<\tau$. We will demonstrate that $\tau = 2h(K,\gamma',T_0)+2a\log (2a)$ satisfies this condition. Using basic algebra,
     \begin{equation*}
         \begin{aligned}
             &h(K,\gamma',T_0)+a\log \tau \\
             & = h(K,\gamma',T_0)+a\log \left(2h(K,\gamma',T_0)+2a\log (2a)\right)\\ 
             &= h(K,\gamma',T_0)+a\log \left(2a\log (2a)\right)+a\log \left(1+\frac{2h(K,\gamma',T_0)}{2a\log (2a)}\right)\\
             &\leq h(K,\gamma',T_0)+a\log \left(2a\log (2a)\right)+a\cdot \frac{2h(K,\gamma',T_0)}{2a\log (2a)} \\
             &\leq h(K,\gamma',T_0)+a\log \left((2a)^{2}\right)+h(K,\gamma',T_0)\\
             & = 2h(K,\gamma',T_0)+2a\log (2a) =\tau~. 
         \end{aligned}
     \end{equation*}
 \end{proof}
 Before presenting the regret bound in its most general form, we introduce the following term.
     \begin{equation*}
         \tilde{C}_\epsilon: = \max\left(T_\epsilon,e^{\frac{1}\epsilon}\right)+\frac{2\alpha \eta^{-1/2}-2-\epsilon}{2\alpha \eta^{-1/2}-1-\epsilon}\left(\frac{MK^2\left(3+2\log \left(d_{\max}^{\gamma}+1\right)\right)}{\min\{1,2\alpha \eta^{-1/2}-1\}\delta \log \eta}\right)^{\frac{1}{2\alpha \eta^{-1/2}-1-\epsilon}}~.
     \end{equation*}
 \begin{theorem}
	\label{theorem: MP complicated}
     For any $\epsilon>0$, $\gamma'\leq \gamma$ and $\delta>0$, denote $\Gamma(m,\gamma')$ as the size of the largest clique player $m$ belongs to in $\mathcal{G}_{\gamma'}$, and define,
     \begin{equation*}
     \begin{aligned}
         \hat{C}_\epsilon(\delta): &= \Delta_{1\max}\sum_{m=1}^{M}\min_{\gamma'\leq \gamma}\left(K^2(\gamma'+2)+\frac{2\mathcal{D}}{\Gamma(m,\gamma')}\log (\frac{2\mathcal{D}}{\Gamma(m,\gamma')})\right)\\
         &+(2C_{\epsilon}(\delta)+K(3\gamma+2))M\Delta_{1\max}~,
     \end{aligned}
     \end{equation*}
     and,
     \begin{equation*}
     \begin{aligned}
         \hat{C}_\epsilon: &= \Delta_{1\max}\sum_{m=1}^{M}\min_{\gamma'\leq \gamma}\left(K^2(\gamma'+2)+\frac{2\mathcal{D}}{\Gamma(m,\gamma')}\log (\frac{2\mathcal{D}}{\Gamma(m,\gamma')})\right)\\
         &+(2\tilde{C}_{\epsilon}+K(3\gamma+2))M\Delta_{1\max}~.
     \end{aligned}
     \end{equation*}
     Let $\chi(\mathcal{G}_{\gamma})$ denote the clique covering number of $\mathcal{G}_{\gamma}$. Then the following holds for Algorithm~\ref{alg:MPRUCB}.
     \begin{enumerate}[label=(\alph*)]
         \item For any $\alpha$ such that $\alpha> 0.5\sqrt{\eta}(1+\epsilon)$ and all times $T$, with probability larger than $1-\delta$,
            \begin{equation*}
               \mathcal{R}(T) \leq \sum_{j>1}\frac{2\alpha \chi(\mathcal{G}_{\gamma})}{\Delta_{1j}}\log T+\hat{C}_\epsilon(\delta)
            \end{equation*}
        \item For any $\alpha$ such that $\alpha> 0.5\sqrt{\eta}(2+\epsilon)$ and all times $T$,
            \begin{equation*}
            \begin{aligned}
                \mathbb{E}\mathcal{R}(T) &\leq \sum_{j>1}\frac{2\alpha \chi(\mathcal{G}_{\gamma})}{\Delta_{1j}}\log T+\hat{C}_\epsilon~.
            \end{aligned}
            \end{equation*}
     \end{enumerate}
 \end{theorem}
\begin{proof}
     First, we observe the following fact. For $t> C_{\epsilon}(\delta)$ with probability larger than $1-\delta$, no pair $(i,i) \neq (1,1)$ can be drawn by any player. Since every player $m$ recommends an arm $i$ iff $i_m(t)=j_m(t)=i$, messages received at rounds $t> C_{\epsilon}(\delta)+\gamma$ contain $\mathcal{B}^{m}_{\mathrm{rec}}\in \{\emptyset,1\}$.  
     
     For any $\gamma'(m) \leq \gamma$ denote by $C_{\gamma'(m)}^{m}$ as a maximal clique in $\mathcal{G}_{\gamma'(m)}$ that player $m$ belongs to with the maximal size, and denote its size as $\Gamma(m,\gamma'(m))$.    
     The proof for the first part relies on the following observation. Lemma~\ref{lemma:MP visitation bound complicated} implies that with probability $1-\delta$, for all such cliques and all players $m$, times $t>T_0=C_\epsilon(\delta)+\gamma$ and arm pairs $(i,j) \neq (1,1)$,
    \begin{equation*}
        \sum_{m \in C_{\gamma'(m)}^{m}}N_{ij}^{m,T_0}(t)\leq D_{ij}\log t+(\gamma'(m) +2)\Gamma(m,\gamma'(m))~.
    \end{equation*}
    Let $\hat{T}^{m}_{\epsilon}(\delta)$ be a round such that $\hat{T}^{m}_{\epsilon}(\delta) > C_{\epsilon}(\delta)+\gamma+\frac{K(K-1)}{2}(\gamma'(m) +2)+\frac{\mathcal{D}}{\Gamma(m,\gamma'(m))}\log \hat{T}^{m}_{\epsilon}(\delta)$. At round $\hat{T}^{m}_{\epsilon}(\delta)$, it holds that,
    \begin{equation*}
    \begin{aligned}
        &\sum_{m \in C_{\gamma'(m)}^{m}}N_{11}^{m,T_0}\left(\hat{T}^{m}_{\epsilon}(\delta)\right) = \Gamma(m,\gamma'(m))(\hat{T}^{m}_{\epsilon}(\delta) - C_\epsilon(\delta)-\gamma)-\sum_{i<j}\sum_{m \in C_{\gamma'(m)}^{m}}N_{ij}^{m,T_0}\left(\hat{T}^{m}_{\epsilon}(\delta)\right)\\
        &\geq \Gamma(m,\gamma'(m))(\hat{T}^{m}_{\epsilon}(\delta) - C_\epsilon(\delta)-\gamma)-\sum_{i<j}D_{ij}\log\hat{T}^{m}_{\epsilon}(\delta) - \frac{K(K-1)}{2}(\gamma'(m) +2)\Gamma(m,\gamma'(m)) \\
        & >0~,
    \end{aligned}
    \end{equation*}
    where we used the fact that at each round there are $\Gamma(m,\gamma'(m))$ draws in the clique $C_{\gamma'(m)}^{m}$. Therefore, for at least one player $m' \in C_{\gamma'(m)}^{m}$ the pair $(1,1)$ is drawn at least once between rounds $C_\epsilon(\delta)+\gamma$ and $\hat{T}^{m}_{\epsilon}(\delta)~$. We now consider two scenarios.
    \begin{itemize}
        \item If $m=m'$, at the round of this first draw, for all $j>1$, we have $u^{m}_{j1}<0.5$, so that arm $1$ is the only arm in the set $\mathcal{C}^{m}$ and $\mathcal{B}^{m}=\{1\}$ for this round regardless of what recommendations arrive.
        \item If $m'\neq m$, a recommendation about arm $1$ arrives to player $m$ before round $\hat{T}^{m}_{\epsilon}(\delta)+\gamma'(m)~$. At that round $\mathcal{B}^{m}_{\mathrm{rec}}\cap \mathcal{C}^{m}=\{1\}$ since a recommendation about no other arm is possible, and so it updates $\mathcal{B}^{m}=1$.
    \end{itemize}
    In both scenarios, player $m$ updates its $\mathcal{B}^{m}$ set to $1$ before round $\hat{T}^{m}_{\epsilon}(\delta)+\gamma'(m)~$. Since arm $1$ remains in $\mathcal{C}^{m}$ for subsequent rounds and no recommendation other than $1$ or $\emptyset$ arrives, the set $\mathcal{B}^{m}$ never changes going forward - so the first arm drawn after round $\hat{T}^{m}_{\epsilon}(\delta)+\gamma'(m)$ by player $m$ is always arm $1$.

    By Lemma~\ref{lemma:MP T_delta definition} it holds that for any $\gamma'(m) \leq \gamma$,
    \begin{equation*}
        \hat{T}^{m}_{\epsilon}(\delta) \leq 2C_{\epsilon}(\delta)+2\gamma+K(K-1)(\gamma'(m) +2)+\frac{2\mathcal{D}}{\Gamma(m,\gamma'(m))}\log (\frac{2\mathcal{D}}{\Gamma(m,\gamma'(m))})~,
    \end{equation*}
    so with probability $1-\delta$ we decompose the regret with respect to the different cliques in $\mathcal{C}_{\gamma}$,
    \begin{equation*}
        \begin{aligned}
            \mathcal{R}(T)&=\sum_{i \leq j}\sum_{m=1}^{M}\frac{\Delta_{1i}+\Delta_{1j}}{2}N_{ij}^{m}(T)\\
            &\leq \sum_{i \leq j}\sum_{m=1}^{M}\frac{\Delta_{1i}+\Delta_{1j}}{2}N_{ij}^{m}(\hat{T}^{m}_{\epsilon}(\delta)+\gamma'(m))\\
            &+\sum_{i \leq j}\sum_{C \in \mathcal{C}_{\gamma}}\sum_{m\in C}\frac{\Delta_{1i}+\Delta_{1j}}{2}N_{ij}^{m,\hat{T}^{m}_{\epsilon}(\delta)+\gamma'(m)}(T)\\
            &\leq \Delta_{1\max}\sum_{m=1}^{M}(\hat{T}^{m}_{\epsilon}(\delta)+\gamma'(m))+\sum_{j>1}\frac{\Delta_{1j}}{2}\sum_{C \in \mathcal{C}_{\gamma}}\sum_{m\in C}N_{ij}^{m,\hat{T}^{m}_{\epsilon}(\delta)+\gamma'(m)}(T)\\
            &\leq \Delta_{1\max}\sum_{m=1}^{M}(\hat{T}^{m}_{\epsilon}(\delta)+\gamma'(m))+\sum_{j>1}\frac{\Delta_{1j}}{2}\sum_{C \in \mathcal{C}_{\gamma}}\sum_{m\in C}N_{ij}^{m,C_{\epsilon}(\delta)}(T)\\
            &\leq \Delta_{1\max}\sum_{m=1}^{M}(\hat{T}^{m}_{\epsilon}(\delta)+\gamma'(m))+\sum_{j>1}\frac{\Delta_{1j}}{2}\sum_{C \in \mathcal{C}_{\gamma}}\left(D_{1j}\log T+(\gamma+2)|C|\right)\\
            &\leq \Delta_{1\max}\sum_{m=1}^{M}\left(K^2(\gamma'(m) +2)+\frac{2\mathcal{D}}{\Gamma(m,\gamma'(m))}\log (\frac{2\mathcal{D}}{\Gamma(m,\gamma'(m))})\right)\\
            &+(2C_{\epsilon}(\delta)+K(3\gamma+2))M\Delta_{1\max}+\sum_{j>1}\frac{\Delta_{1j}}{2}\sum_{C \in \mathcal{C}_{\gamma}}D_{1j}\log T~.
        \end{aligned}
    \end{equation*}
    In the derivation above we were free to select any $\gamma'(m)\leq \gamma$ for each player, so for the final bound we get,
    \begin{equation*}
        \begin{aligned}
            \mathcal{R}(T)&\leq \Delta_{1\max}\sum_{m=1}^{M}\min_{\gamma'\leq \gamma}\left(K^2(\gamma'+2)+\frac{2\mathcal{D}}{\Gamma(m,\gamma')}\log (\frac{2\mathcal{D}}{\Gamma(m,\gamma')})\right)\\
            &+(2C_{\epsilon}(\delta)+K(3\gamma+2))M\Delta_{1\max}+\sum_{j>1}\frac{\Delta_{1j}}{2}\sum_{C \in \mathcal{C}_{\gamma}}D_{1j}\log T\\
            &\leq \hat{C}_{\epsilon}(\delta)+\sum_{j>1}\frac{\Delta_{1j}}{2}\chi(G_{\gamma})D_{1j}\log T~.
        \end{aligned}
    \end{equation*}
    
    To prove the second part, note that the following holds for any random variable X: $\mathbb{E} X = \int_{0}^{1}F_{X}^{-1}(q)dq~$, where $F_X^{-1}(\cdot)$ stands for the quantile function of $X$. For some invertible function $H(q)$, given that $F_{\mathcal{R}(T)}(H(q))=P\left(\mathcal{R}(T) \leq H(q)\right)>q$, it holds that $F_{\mathcal{R}(T)}(r) > H^{-1}(r)$, so $F_{\mathcal{R}(T)}^{-1}(q) < H(q)$. In our case with $q=1-\delta$, $H(q)$ depends on $q$ only through the first term in the regret containing $\hat{C}_\epsilon(1-q)~$. For clarity, we denote,
    \begin{equation*}
       \hat{C}_\epsilon(\delta) = \max\left(A,\left(\frac{B}{\delta}\right)^{\frac{1}{2\alpha \eta^{-1/2}-1-\epsilon}}\right)~,
    \end{equation*}
    where $A,B$ are some terms that are not functions of $\delta$. For $2\alpha \eta^{-1/2} > 2+\epsilon$, it holds that,
    \begin{equation*}
        \begin{aligned}
            \int_{0}^{1}\left(\frac{1}{1-q}\right)^{\frac{1}{2\alpha \eta^{-1/2}-1-\epsilon}}dq=\int_{0}^{1}\left(\frac{1}{\delta}\right)^{\frac{1}{2\alpha \eta^{-1/2}-1-\epsilon}}d\delta = \frac{2\alpha \eta^{-1/2}-2-\epsilon}{2\alpha \eta^{-1/2}-1-\epsilon}~.
        \end{aligned}
    \end{equation*}
    Putting it all together we have that, 
    \begin{equation*}
        \mathbb{E}\mathcal{R}(T) \leq \hat{C}_\epsilon+\sum_{j>1}\frac{\Delta_{1j}}{2}\chi(\mathcal{G}_{\gamma})D_{1j}\log T~.
    \end{equation*}
 \end{proof}
The derivation of the proof for Theorem~\ref{theorem: MP} in the paper follows a similar process, assuming $\alpha>1.4, \epsilon=e^{-1},\eta=1.3$ and utilizing $C(\delta)$. For a graph with complete communication $\gamma=D$, the non-leading term in the regret becomes,
\begin{equation*}
     \begin{aligned}
         \hat{C}: &= \Delta_{1\max}\sum_{m=1}^{M}\min_{\gamma'\leq D}\left(K^2(\gamma'+2)+\frac{2\mathcal{D}}{\Gamma(m,\gamma')}\log 2\mathcal{D}\right)+(2\tilde{C}+K(3D+2))M\Delta_{1\max}~.
     \end{aligned}
 \end{equation*}
Since a minimization is involved in the calculation of this term, we can observe the two extreme cases where $\gamma'=0$,
\begin{equation*}
     \begin{aligned}
         \hat{C}&\leq \Delta_{1\max}\sum_{m=1}^{M}\left(2K^2+2\mathcal{D}\log 2\mathcal{D}\right)+(2\tilde{C}+K(3D+2))M\Delta_{1\max}\\
         &=O(MKD+MK^2\log K/\Delta^2_{1\min})~,
     \end{aligned}
\end{equation*}
and $\gamma'=D$,
\begin{equation*}
     \begin{aligned}
         \hat{C}&\leq \Delta_{1\max}\sum_{m=1}^{M}\left(K^2(D+2)+\frac{2\mathcal{D}}{M}\log 2\mathcal{D}\right)+(2\tilde{C}+K(3D+2))M\Delta_{1\max}~\\
         &=O(MK^2 D+K^2\log K/\Delta^2_{1\min})~.
     \end{aligned}
\end{equation*}
Since both must hold, we have that,
\begin{equation*}
     \begin{aligned}
         \hat{C}=O\left(MKD+\frac{K^2\log K}{\Delta^2_{1\min}}+MK^2\min\Big\{\frac{\log K}{\Delta^2_{1\min}},D\Big\}\right)~.       
     \end{aligned}
\end{equation*}
\section{MORE EXPERIMENTS}
\label{sec: Appendix D: More Experiments}
In this section, we provide some more details about the experimental results. The code used to create the figures in this paper, as well as the Sushi and Irish datasets and all preference matrices, is available at \url{https://github.com/ravehor92/Multi-Player-Approaches-for-Dueling-Bandits}. The Irish and Sushi datasets were obtained using the PrefLib dataset library \citep{MaWa13a}. Please make sure to read the 'README.md' file before accessing the code. For the figures in the main paper, we use $\alpha=3$ for RUCB as both the base algorithm for Algorithm~\ref{alg:FYLBlacBox} and in Algorithm~\ref{alg:MPRUCB}.
Regarding RMED2FH, we use $\alpha=3$ and $f(K)=0.3K^{1.01}$ following \citet{komiyama2015regret}. For FYLBB algorithms, we utilize the simple leader election algorithm from Appendix~\ref{subsec: Leader Election Algorithm}. The VDB algorithm is run with a time-dependent learning rate $\eta_t = 4/\sqrt{t}$ as indicated in \citet{saha2022versatile}, and the emergent optimization problem at each round is solved using the cvxpy library in Python. All regrets are averaged over 200 independent runs, and the error bars indicate the standard deviation over the runs as calculated using numpy.std. Figures in the main paper use $\gamma= D$. Additionally, the horizontal axis in all plots is scaled logarithmically with $T$ to highlight the $\log T$ dependence of the regret in the experiments. The vertical axis is either linear or logarithmic, chosen to improve plot visibility. For consistency, subfigures within the same figure share identical axis properties.

Our experiments were conducted using Python 3.6 on a system featuring a 2.2 GHz Intel Xeon E5-2698 v4 CPU with 15 cores and 512GB of DDR4 RAM operating at 2133 MHz. All experiments were completed within several hours. 

The large variance in Figure \ref{fig: Different Graphs}c originates from the RMED2FH algorithm used by the leader, as asymptotically optimal algorithms explore less and tend to have larger variance. In addition, since the exploration in Figure \ref{fig: Different Graphs}c originates solely from the leader and the followers exploit its candidate arm, this effect is amplified when there are more players. In contrast, exploration in Figure \ref{fig: Different Graphs}a is done cooperatively so the variance is reduced.

\begin{figure}[ht]
\vskip 0.2in
\begin{center}
\centerline{\includegraphics[width=0.6\textwidth]{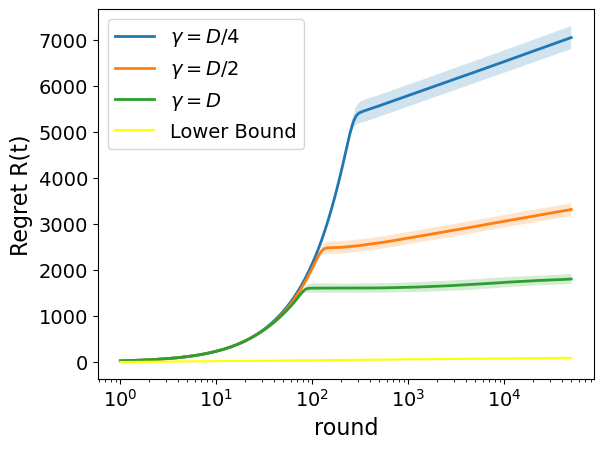}}
\caption{Regret for the Sushi dataset, with the message passing RUCB algorithm and M=100 players. We use a cyclic communication graph and various values for the decay parameter $\gamma$.}
\label{fig: connectivity}
\end{center}
\vskip -0.2in
\end{figure}
Figure~\ref{fig: connectivity} illustrates the group regret per round for the Sushi dataset, employing the message-passing RUCB algorithm with 100 players on a cyclic communication graph. The parameter $\gamma$ is varied to observe its impact on the regret. As $\gamma$ decreases, the slope in the asymptotic region increases, and notably, the case where $\gamma=D$ aligns its slope closest to the lower bound. This alignment corresponds consistently with the findings outlined in Theorem~\ref{theorem: MP}.

\begin{figure}[H]
\vskip 0.2in
\begin{center}
\centerline{\includegraphics[width=0.6\textwidth]{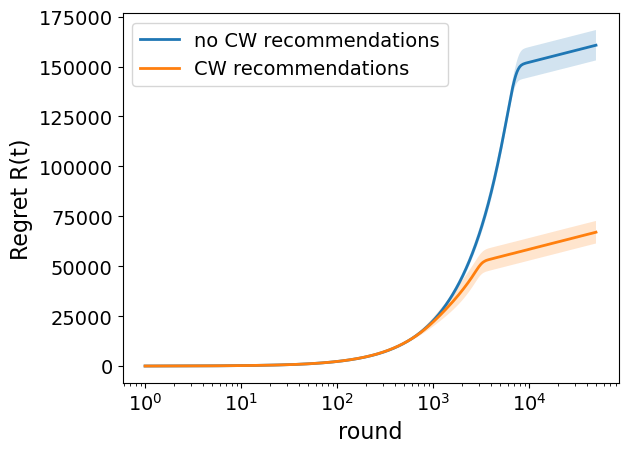}}
\caption{Regret for the Sushi dataset, with M=100 players on a star graph and $\gamma=1$. We use the message passing RUCB algorithm from Section~\ref{sec: A Fully Distributed Approach}, as well as a version without CW recommendations.}
\label{fig: star}
\end{center}
\vskip -0.2in
\end{figure}

In Figure~\ref{fig: star}, we present the regret per round for the Sushi dataset, with M=100 players arranged on a star graph. We compare the performance of the Message Passing RUCB algorithm, as described in Section~\ref{sec: A Fully Distributed Approach}, with and without CW recommendations. Setting the delay to $\gamma=1$ allows us to illustrate the advantage of CW recommendations in quickly identifying the CW. In this scenario, the central player has immediate access to all other players' observations, while the peripheral players can only utilize their own observations and those of the central player. Consequently, we anticipate that the central player will identify the CW faster than the peripheral players. Once this occurs a CW recommendation will benefit all peripheral players by helping them identify the CW and focus on exploring pairs of arms where at least one arm is the CW. This approach helps address the inherent challenge in the multi-player dueling bandit setting, as discussed in Section~\ref{sec: Introduction}, which involves exploring non-CW pairs of arms.
Both versions of the Message Passing UCB exhibit similar asymptotic performance. However, as shown in Figure~\ref{fig: star}, utilizing CW recommendations leads to a quicker identification of the CW and subsequently results in a much smaller regret. This effect is particularly pronounced in experimental settings where communication graphs have $\gamma < D$.


\end{document}